\documentclass{article}

 \usepackage[main, final, nonatbib]{neurips_2026}

\usepackage[utf8]{inputenc} 
\usepackage[T1]{fontenc}    
\usepackage{comment}
\usepackage{graphicx}
\usepackage{wrapfig}
\usepackage{svg}
\usepackage{subcaption}
\usepackage{amsmath}
\usepackage{xcolor}
\usepackage{tikz}

\definecolor{takeawayblue}{rgb}{0.403440,0.624540,0.826370}
\newcommand{\takeaway}[1]{%
    \par\begingroup
    \noindent\begin{tikzpicture}
        \node[draw=none, fill=takeawayblue!30!white,
            rounded corners=8pt, line width=0pt,
            inner sep=6pt, outer sep=0pt,
            text width=\dimexpr\linewidth-12pt\relax,
            align=justify, text=black] {\textbf{Takeaway.}\enspace\textit{#1}};
    \end{tikzpicture}\par
    \endgroup
}

\usepackage{booktabs}
\usepackage{pifont}
\usepackage{makecell}
\usepackage{multirow}

\usepackage[
    backend=biber,
    style=numeric,
    citestyle=numeric-comp,
    sorting=none,
    maxbibnames=5,
    minbibnames=1
]{biblatex}
\usepackage{hyperref}       
\definecolor{managuagold}{rgb}{1.0,0.81263,0.40424}
\definecolor{managuablue}{HTML}{608EC7}
\definecolor{managuaorange}{HTML}{E9AC5B}
\definecolor{managualink}{rgb}{0.756470,0.456630,0.285610}
\hypersetup{
    colorlinks=true,
    allcolors=managualink
}
\usepackage{url}            
\usepackage{booktabs}       
\usepackage{amsfonts}       
\usepackage{nicefrac}       
\usepackage{microtype}      
\usepackage{xcolor}         

\usepackage[capitalize]{cleveref}

\title{Understanding Multi-View Transformers}

\author{
  \normalfont
  \begin{tabular}{c}
    \textbf{Julien Gaubil}\textsuperscript{1}\thanks{Equal contribution. Work partially done while visiting students at MIT.} \quad
    \textbf{Michal Star\'{y}}\textsuperscript{2 \ensuremath{{*}}} \quad
    \textbf{Louis Martinez}\textsuperscript{3} \\
    \textbf{Andreas Geiger}\textsuperscript{4, 5} \quad
    \textbf{Maks Ovsjanikov}\textsuperscript{1}\thanks{Equal supervision.} \quad
    \textbf{Ayush Tewari}\textsuperscript{6 \ensuremath{\dagger}} \quad
    \textbf{Vincent Sitzmann}\textsuperscript{7 \ensuremath{\dagger}} \\[0.35em]
    \textsuperscript{1}LIX, École Polytechnique, IPP \quad
    \textsuperscript{2}TUM \quad
    \textsuperscript{3}Lamarck Labs \\
    \textsuperscript{4}KE:SAI \quad
    \textsuperscript{5}T\"ubingen AI Center, University of T\"ubingen \\
    \textsuperscript{6}University of Cambridge \quad
    \textsuperscript{7}MIT CSAIL
  \end{tabular}
}

\begin{document}

\maketitle

\begin{figure}[ht!]
    \centering
    \includegraphics[width=\textwidth]{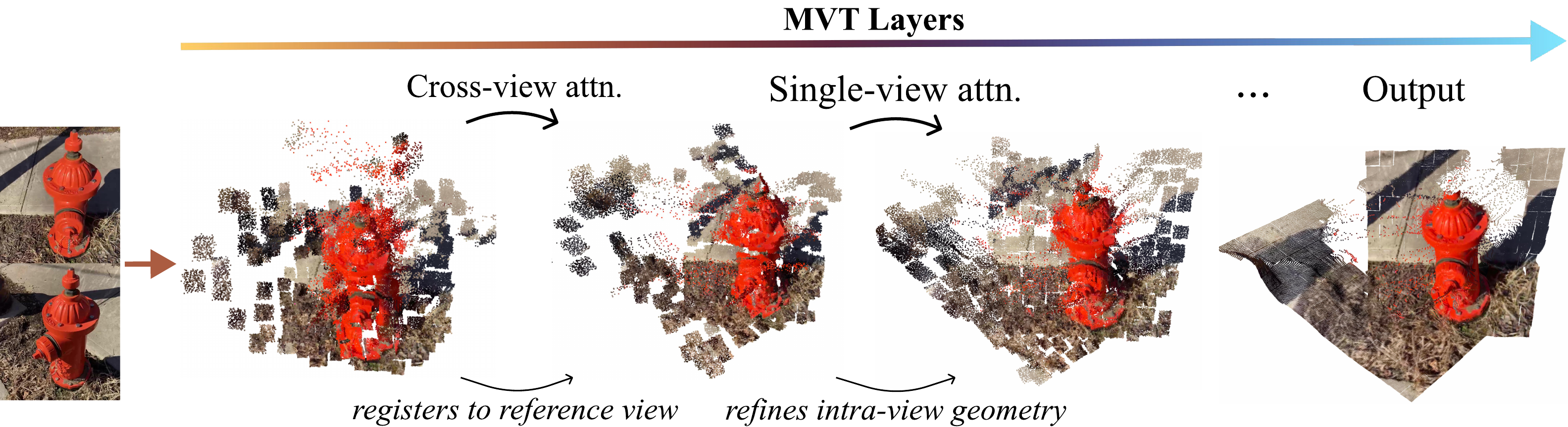}
   \caption{\textbf{Probing multi-view transformers.} We train layer-wise pointmap probes which decode intermediate tokens into dense 3D reconstructions and demonstrate that multi-view transformers reconstruct 3D scenes through progressive refinement. Across layers, cross-view attention improves multi-view consistency, while single-view attention refines geometry within the non-reference views.}
    \label{fig:teaser}
\end{figure}

\begin{abstract}


Multi-view transformers solve 3D vision tasks in a single feed-forward pass, yet their internal mechanisms remain poorly understood. We view their residual layers as progressively updating a latent state, and study how this state evolves during 3D reconstruction. To make the state geometrically interpretable, we train layer-wise probes that decode intermediate tokens into 3D pointmaps, yielding a 3D reconstruction at every layer of the network. We show that multi-view transformers reconstruct 3D scenes progressively, with most layers refining the reconstruction, even though they operate in a single forward pass. Reconstruction speed varies with sample difficulty: easy samples converge within a few layers, whereas difficult ones require more updates. We analyze these updates and find consistent, complementary roles for single-view and cross-view attention. We demonstrate that cross-view attention improves multi-view alignment, while single-view attention refines intra-view geometry. Finally, we identify cross-view attention heads that specialize in correspondence matching. We find that correspondences produced by these heads are refined jointly with the reconstructed geometry, and that encoder choice strongly affects correspondence accuracy. Overall, our analysis provides a direct view of the reconstruction process inside multi-view transformers, and suggests similarities between these feed-forward models and iterative Structure-from-Motion pipelines. Visualizations, code, and probe checkpoints are available on our \href{https://jgaubil.com/research/understanding-mvt}{project page}.

\end{abstract}

\section{Introduction}
\label{sec:intro}

Reconstructing 3D scenes from images has traditionally relied on explicit, iterative optimization pipelines such as COLMAP~\cite{schoenberger2016sfm}. Models in this family estimate correspondences, camera poses, and scene geometry through a structured sequence of updates. Recent multi-view transformers (MVTs) depart from this paradigm. Starting with DUSt3R~\cite{dust3r_cvpr24}, such approaches replace this expert-designed pipeline with a generic transformer that processes two or more views and predicts scene geometry in a single forward pass. This design has rapidly become a leading approach for 3D reconstruction~\cite{wang2025vggt,jin2024stereo4d,zhang2024monst3r,wang2025pi,keetha2026mapanything,depthanything3} and 3D-aware tasks such as novel view synthesis~\cite{srt22,sajjadi2022rust,jiang2025rayzer,jin2025lvsm,mitchelryu2025xfactor,kim2026svsm} or correspondence matching~\cite{mast3r_arxiv24}. Compared to conventional optimization-based methods, feed-forward models are fast and can solve underdetermined reconstruction problems. Yet their internal computation remains poorly understood, beyond evidence that geometric structures such as correspondences~\cite{an2024zeroco} and the fundamental matrix~\cite{bratulic2025undvggt} can emerge in their representations. Understanding this computation can inform useful inductive biases~\cite{miyato2024gta,li2025cameras}, or support zero-shot extensions to dynamic scenes~\cite{chen2025easi3r,han2025d}.

Prior work has studied residual networks --- transformers in particular --- through an iterative lens, casting inference as a process that carries and updates a state through each layer's skip connection~\cite{alain2018probing,jastrzębski2018residual,geva2022transformer,dar2023analyzing,molina2023traveling,belrose2023elicitinglatentpredictionstransformers,elhage2021mathematical}. Inspired by this view, we study how spatial geometry emerges in the residual state of multi-view transformers. Specifically, we train small probes~\cite{alain2018probing} to decode individual patch tokens into explicit 3D pointmaps after each residual update. Unlike transformer analyses in vocabulary or language embedding space~\cite{geva2022transformer,dar2023analyzing,molina2023traveling,belrose2023elicitinglatentpredictionstransformers}, 3D has a well-studied geometric structure that lets us inspect the layer-by-layer sequence of decoded states. Moreover, by probing multi-view transformers on their native reconstruction task, we directly study how these models solve multi-view 3D reconstruction.

Using this framework, we examine recent multi-view transformers: DUSt3R~\cite{dust3r_cvpr24}, VGGT~\cite{wang2025vggt}, and Depth Anything 3~\cite{depthanything3}. We demonstrate that, although they reconstruct 3D scenes in a single forward pass, their intermediate reconstructions \textit{gradually} move toward the final prediction. Most layers improve the reconstruction on average. Beyond identifying the geometric information present in intermediate features, our probes reveal \textit{how} this information changes across layers. We find consistent and complementary roles for single-view and cross-view attention. Cross-view attention increases multi-view consistency by aligning predictions to the reference view, while single-view attention improves intra-view geometry in non-reference frames. Together, these updates progressively register the views into a locally and globally coherent reconstruction. Finally, we show that correspondences emerging in MVT cross-view attention maps~\cite{an2024zeroco,bratulic2025undvggt} are refined jointly with internal geometry.

Overall, multi-view transformers trained on large 3D datasets show similarities to traditional pipelines such as Structure-from-Motion~\cite{schoenberger2016sfm}: network depth acts as a refinement budget, geometry progressively improves, and correspondences emerge and are used during reconstruction. We further retrain MVTs from scratch in a unified codebase to study how design affects these properties. In small-scale experiments with matched compute budgets and parameter counts, attention-only MVTs outperform MVTs that include MLP layers, which lack a clear geometric interpretation in our framework. Additionally, models trained with CroCoV2~\cite{croco_v2} encoders exhibit significantly higher correspondence accuracy in their cross-view attention maps than models trained with DINOv2~\cite{oquab2023dinov2} encoders.

Taken together, our work provides a systematic picture of the learned reconstruction process implemented by multi-view transformers. Our contributions are \textbf{1)} a layer-wise pointmap probing framework applied to study the 3D reconstruction process of several recent multi-view transformers; \textbf{2)} evidence that feed-forward reconstruction is internally progressive, with complementary roles for single-view and cross-view attention in this refinement; and \textbf{3)} an analysis of the correspondences that emerge in cross-view attention maps and are refined across layers alongside the geometry.
\begin{table}[!t]
\centering
\caption{\textbf{Design choices of the studied multi-view transformers.} Models differ in encoder and MVT initialization, cross-view attention parameterization (CVA), and output 3D parameterization.}
\begin{tabular}{
    l
    >{\centering\arraybackslash}p{2.15cm}
    >{\centering\arraybackslash}p{3.3cm}
    >{\centering\arraybackslash}p{4.4cm}
}
\toprule
& \textbf{DUSt3R}~\cite{dust3r_cvpr24}
& \textbf{DA3}~\cite{depthanything3}
& \textbf{VGGT}~\cite{wang2025vggt} \\
\midrule
\textbf{Encoder init.} & CroCoV2~\cite{croco_v2} & DINOv2~\cite{oquab2023dinov2}$^\dagger$ & DINOv2~\cite{oquab2023dinov2} \\
\textbf{MVT init.} & CroCoV2~\cite{croco_v2} & DINOv2~\cite{oquab2023dinov2}$^\dagger$ & Random \\
\textbf{MVT weights} & Per-view & View-shared & View-shared \\
\textbf{CVA param.} & Cross-attention & Global self-attention & Global self-attention \\
\textbf{Output param.} & Pointmap & Depth + pose + ray map & Pointmap + depth + pose + FoV \\
\bottomrule
\end{tabular}

\vspace{0.25em}
\footnotesize\raggedright
$^\dagger$ DA3 initializes its encoder and MVT using the first third and remaining two thirds of DINOv2, respectively.
\label{tab:model_comparison}
\end{table}

\section{Related Work and Preliminaries}
\label{sec:related}

\paragraph{3D reconstruction from images.}
Reconstructing 3D scenes from images is a long-standing problem~\cite{Hartley2003mvg}. Classical Structure-from-Motion (SfM) systems such as COLMAP~\cite{schoenberger2016sfm} solve unposed reconstruction through an explicit pipeline of correspondence matching, relative pose estimation, triangulation, and joint optimization of camera poses and geometry. Learning-based methods have replaced or augmented parts of this pipeline with learned components~\cite{smith24flowmap,brachmann2024acezero,wang2024vggsfm}. DUSt3R~\cite{dust3r_cvpr24} marks a more direct departure: it uses a vision transformer (ViT)~\cite{vasnawi2017attention,dosovitskiy2021an} to jointly process two views and predict scene geometry in a single forward pass. This \textit{multi-view transformer} design quickly became a de facto choice for 3D reconstruction tasks~\cite{wang2025vggt,jin2024stereo4d,zhang2024monst3r,wang2025pi,keetha2026mapanything,depthanything3,tang2024mv,Yang_2025_Fast3R,jang2025pow3r}. Other approaches process views sequentially while updating a latent spatial memory~\cite{wang2024spann3r,cut3r,must3r_cvpr25}, which naturally supports long image sequences. Replacing handcrafted SfM updates with generic attention blocks improves speed and robustness, but makes reconstruction less transparent. In this paper, we show that multi-view transformers learn interpretable operations that resemble traditional optimization pipelines.

\paragraph{Multi-view transformers.}
\begin{wrapfigure}[17]{r}{0.48\textwidth}
    \vspace{-1.5em}
    \centering
    \captionsetup[subfigure]{justification=centering}
    \begin{subfigure}[t]{0.5548\linewidth}
        \centering
        \includegraphics[width=\linewidth]{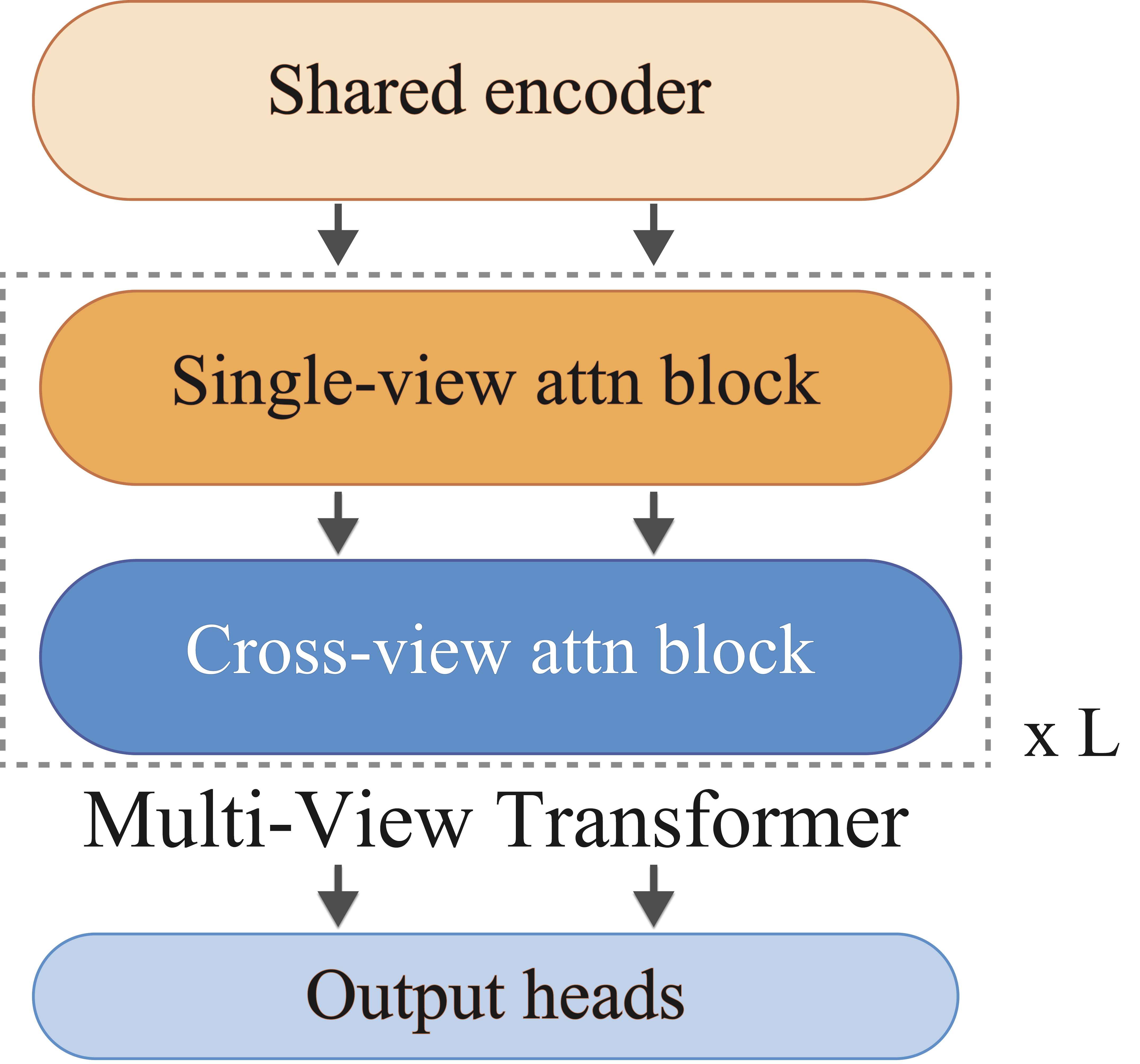}
        \caption{MVT overview}
        \label{fig:overview-mvt}
    \end{subfigure}\hfill
    \begin{subfigure}[t]{0.44\linewidth}
        \centering
        \includegraphics[width=0.8300\linewidth]{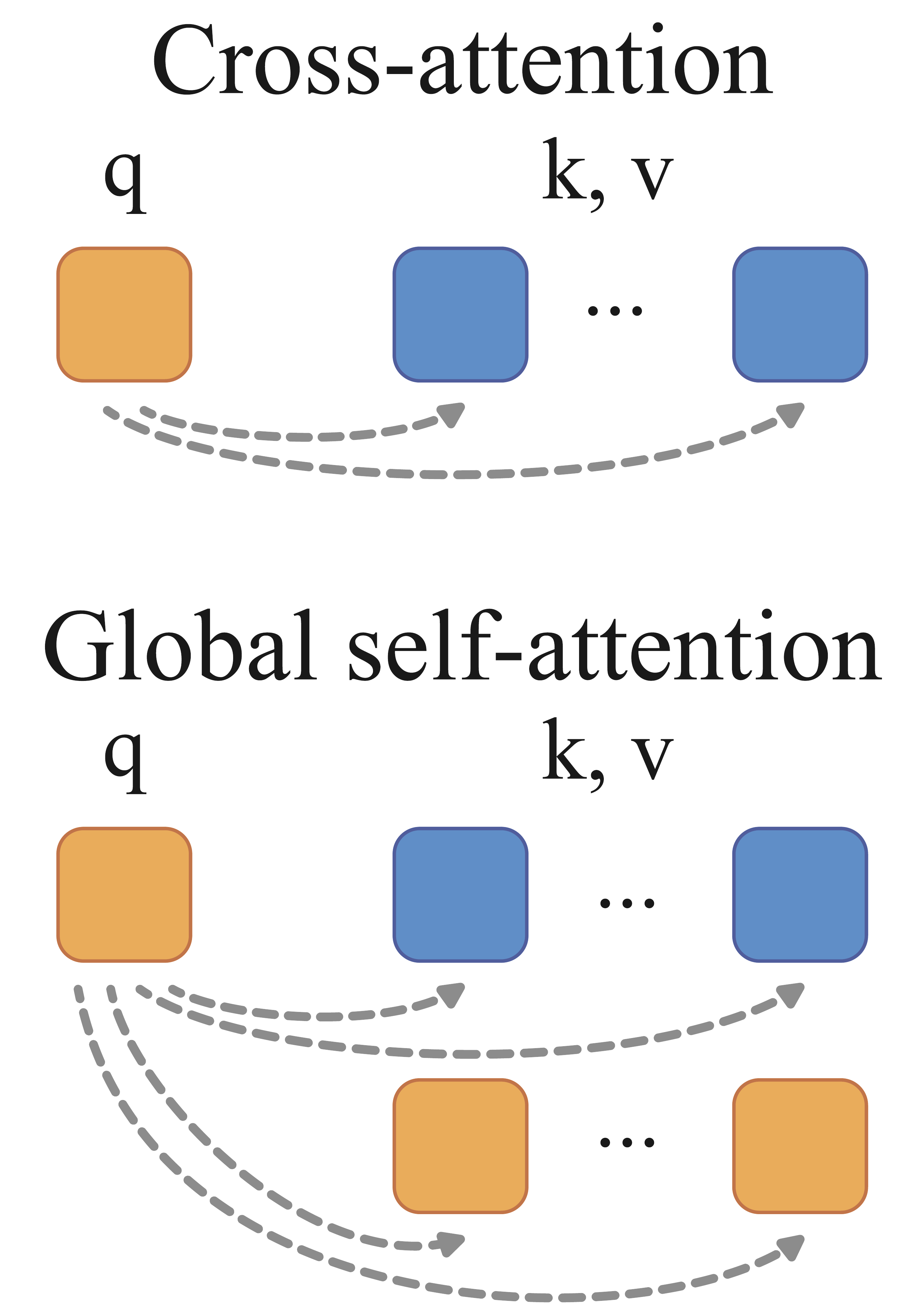}
        \caption{CVA mechanisms}
        \label{fig:overview-cva}
    \end{subfigure}
    \caption{\textbf{Multi-view transformers overview.}
    \textbf{(a)} MVTs process all frames jointly following a monocular image encoder. \textbf{(b)} Cross-view attention (CVA) can be implemented with cross-attention to tokens from \textcolor{managuablue}{other views}, or global self-attention over tokens from \textcolor{managuablue}{all} \textcolor{managuaorange}{views}.
    }
    \vspace{-1em}
    \label{fig:overview}
\end{wrapfigure}
Beyond reconstruction, multi-view transformers have been adopted for 3D-aware tasks such as correspondence matching~\cite{mast3r_arxiv24} and novel view synthesis~\cite{srt22,sajjadi2022rust,jiang2025rayzer,jin2025lvsm,mitchelryu2025xfactor,kim2026svsm}. Their architecture is illustrated in~\cref{fig:overview-mvt}: a shared encoder first processes each view separately, then a multi-view transformer jointly processes all views before task-specific heads predict the final outputs. This design has several advantages: it can reuse pretrained ViT representations, supports various forms of cross-view communication through attention (see~\cref{fig:overview-cva}), and adapts to many downstream tasks. Our study focuses on DUSt3R, VGGT, and Depth Anything 3 (DA3)~\cite{dust3r_cvpr24,wang2025vggt,depthanything3}, whose main design differences in terms of initialization, cross-view attention mechanism, and output parameterization are summarized in~\cref{tab:model_comparison}.

\paragraph{Analyzing iterative inference in transformers.}
Residual networks~\cite{He_2016_CVPR} can be interpreted as iterative inference systems: each residual layer updates a latent state, and training encourages these updates to reduce the loss~\cite{jastrzębski2018residual}. This view has been used to study how transformer representations evolve across depth in language models~\cite{geva2022transformer,dar2023analyzing,molina2023traveling,belrose2023elicitinglatentpredictionstransformers}. Such analyses often rely on probing~\cite{alain2018probing}, which makes intermediate states inspectable by training lightweight heads on frozen representations. In 3D vision, prior probing work has mostly studied monocular encoders at one or a few layers~\cite{elbanani2024probing,man2024lexicon3d,zhan2024physical,danier2025depthcues,bonnen2024mochi,michalkiewicz2025giq,chen2025feat2gs}. We instead adopt the iterative perspective and probe the multi-view transformer itself, layer by layer, to analyze how it exchanges information across views and progressively reconstructs 3D geometry.

\paragraph{Understanding and improving multi-view transformers.}
Existing analyses of multi-view transformers mostly focus on attention maps~\cite{an2024zeroco,han2025d,chen2025easi3r,han2025emergent}. Closest to our work, Bratuli{\'c} et al.~\cite{bratulic2025undvggt} probe MVT layers for the fundamental matrix and study the role of correspondences and learned priors in reconstruction. We instead probe for 3D reconstructions directly from intermediate states. This reveals how multi-view transformers progressively refine geometry despite operating in a single forward pass, and clarifies each layer's role in the reconstruction process.
\begin{figure}[!t]
    \centering
    \begin{subfigure}[b]{0.3916\linewidth}
        \centering
        \includegraphics[width=\linewidth]{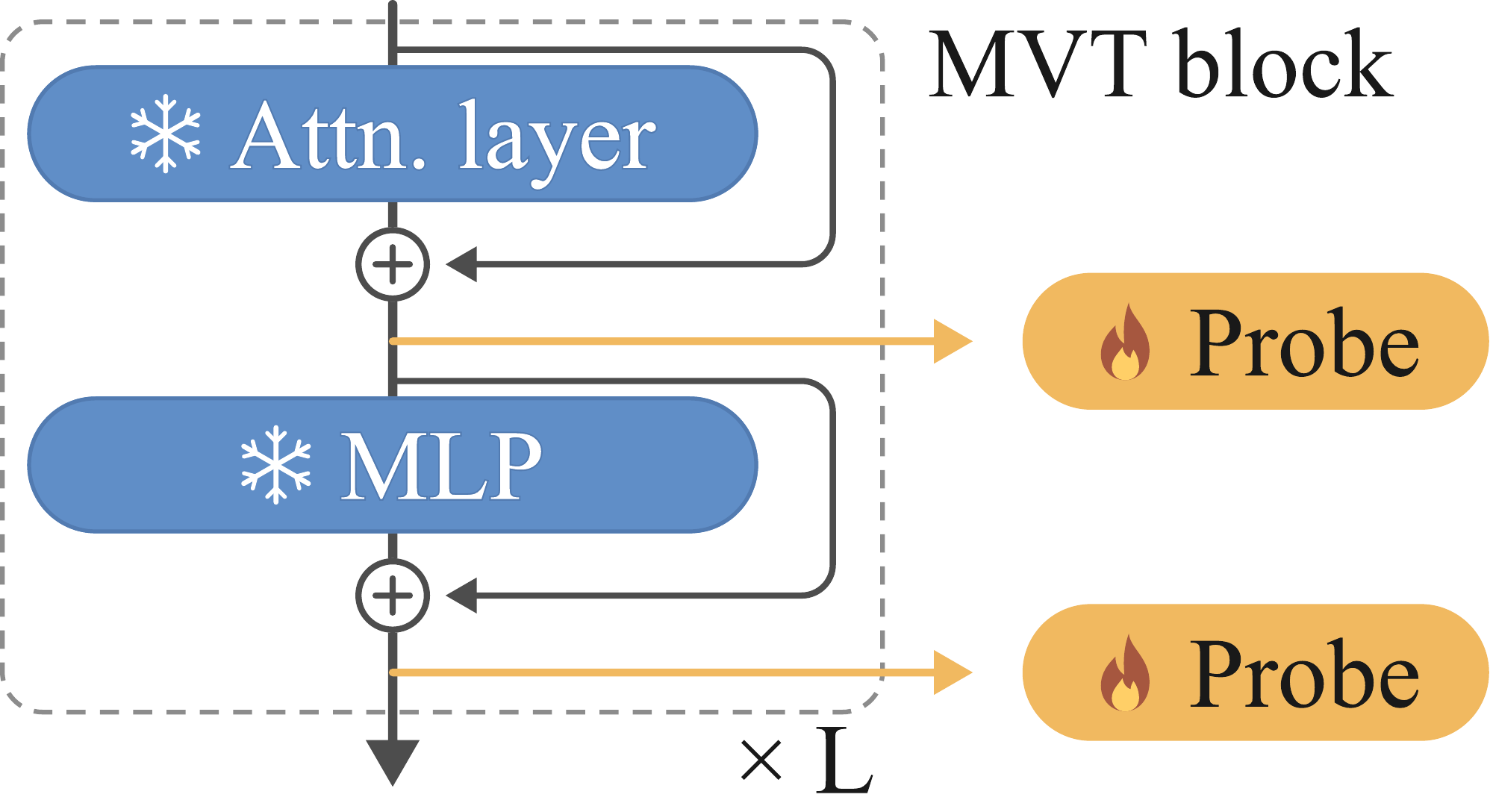}
        \caption{Probe placement}
        \label{fig:probe-states}
    \end{subfigure}\hfill
    \begin{subfigure}[b]{0.5584\linewidth}
        \centering
        \includegraphics[width=\linewidth]{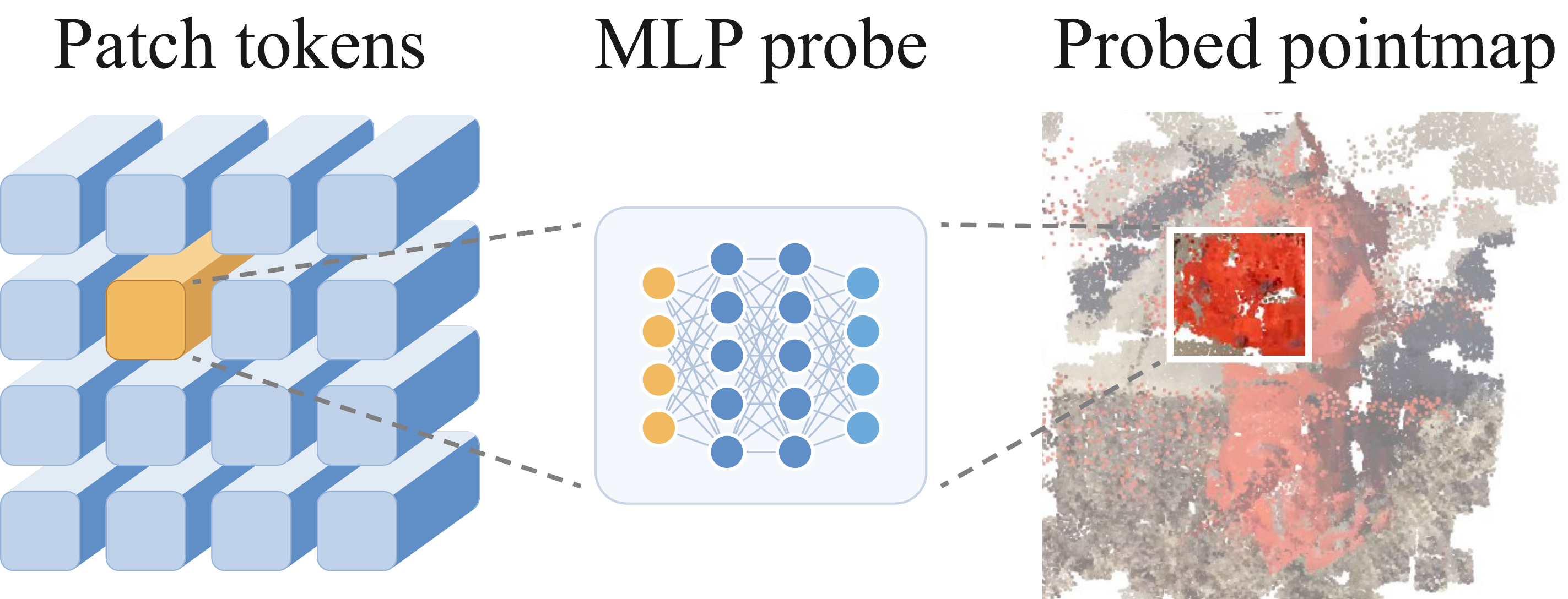}
        \caption{Probe mechanism}
        \label{fig:probe-mechanism}
    \end{subfigure}
    \caption{\textbf{Probing mechanism.} \textbf{(a)} We train per-layer probes after the residual update within frozen multi-view transformer blocks. \textbf{(b)} The probes map individual patch tokens to their corresponding 3D pointmap patches. Probes are separate across layers but shared across patches within a layer.}
    \label{fig:probe}
    \vspace{1mm}
\end{figure}

\section{Understanding Multi-View Transformers}
\label{sec:method}

\subsection{Probing Residual Pointmaps}\label{subsec:pointmap-probe}

To study how geometry evolves in the residual state of a multi-view transformer, we train probes on patch features extracted after each residual update.

\paragraph{Probe 3D parameterization.}
Our probes predict pointmaps: per-pixel 3D points expressed in a coordinate frame shared across views. Unlike local parameterizations such as depth, pointmaps place all views at a common scale and pose, facilitating joint visualization and evaluation. This makes them well suited for inspecting the 3D structure encoded in the latent state. We express all pointmaps in the coordinate frame of the first, reference view, mirroring the training setup of the evaluated models.

\paragraph{Multi-view setup.}
We mainly focus on two-view reconstruction to study how MVTs solve this central multi-view geometry problem. Two-view reconstruction can be extended to \(n\) views with iterative registration procedures, and it lets us compare DUSt3R, which takes two views as input, with VGGT and DA3, which can process longer sequences. For VGGT, we conduct additional four-view experiments to study feed-forward reconstruction beyond the two-view case.

\paragraph{Probe design and training.}
We train two probes, one for the reference view and one shared across all remaining views. This reflects the asymmetry of the prediction tasks: reconstructing the reference-view pointmap requires estimating depth and intrinsics, while reconstructing a non-reference-view pointmap also requires estimating that view's pose relative to the reference. We limit the probe capacity to ensure that it faithfully reflects the 3D information encoded in the state: each probe maps individual patch features to 3D points without communication across patches. This prevents the probe from solving global reasoning tasks on its own, such as relative pose estimation or joint geometry reconstruction. We also regress per-view confidence maps and train probes with the confidence-aware loss from DUSt3R~\cite{dust3r_cvpr24}. \cref{fig:probe} illustrates the probe mechanism and Appendix~\ref{app:train_details} details the loss.

\subsection{Evaluating Pointmap Probes}\label{subsec:pointmap-eval}

\paragraph{Pointmap metrics.}
We evaluate predicted pointmaps with three errors. First, \(e_{\mathrm{point}}\) is the absolute-relative 3D pointmap error used in~\cite{keetha2026mapanything}, computed after scale normalization. We also report a rigidly aligned variant, \(e_{\mathrm{proc}}\), obtained by applying a per-view Procrustes alignment between predicted and ground-truth pointmaps before computing the same error. This metric evaluates intra-view geometry independently of relative pose error. Finally, \(e_{\mathrm{mv}}\) measures multi-view 3D consistency as the \(e_{\mathrm{point}}\) error between predictions for corresponding 3D points in different views. Metrics are computed per sample and view, then averaged over test samples.

\paragraph{Difficulty subsets.}
We categorize test samples by reconstruction difficulty. For each model, we compute \(e_{\mathrm{point}}\) for the final pretrained reconstruction and obtain one error score per sample. We then define per-model easy and hard subsets as the bottom and top 20\% of samples, respectively.

\paragraph{Convergence analysis.}
To understand how MVT layers progressively refine the reconstruction toward a final estimate, we study how intermediate probe reconstructions approach the final probe reconstruction across depth. We measure \(e_{\mathrm{point}}\) between each intermediate pointmap and the final probe reconstruction, rather than between the intermediate pointmap and ground truth. This yields per-layer error sequences that approach zero regardless of the final reconstruction error, allowing comparison across difficulty subsets and models. We summarize convergence with the inlier ratio at an \(e_{\mathrm{point}}\) threshold of \(0.05\), averaged across samples.

\paragraph{Measuring layer contributions.}
We quantify how different MVT layer types contribute to reducing a pointmap error \(e\). Let \(e_k\) denote the dataset-averaged value of \(e\) for pointmaps probed after layer \(k\), and let \(\mathcal{K}_{\mathcal{F}}\) be the set of layer indices with type \(\mathcal{F}\), where \(\mathcal{F}\) is single-view attention, cross-view attention, or MLP. We define the Accumulated Layer Contribution Percentage (ALCP) as:

\begin{equation}\label{eq:ALCP}
    \mathrm{ALCP}(e, \mathcal{F}) = 100 \cdot \frac{\sum_{k \in \mathcal{K}_{\mathcal{F}}} \left(e_{k-1} - e_k\right)}{e_{\mathrm{enc}} - e_{\mathrm{mvt}}},
\end{equation}
where \(e_{\mathrm{enc}}\) and \(e_{\mathrm{mvt}}\) are the dataset-averaged errors \(e\) at the encoder and MVT output, respectively. A positive \(\mathrm{ALCP}(e, \mathcal{F})\) indicates that layers of type \(\mathcal{F}\) reduce the error \(e\) on average across the MVT.

We detail metrics, alignment procedures, and difficulty subset definitions in Appendix~\ref{subsec:app-pointmap_eval_details}.

\subsection{Additional Zero-shot Evaluations}\label{subsec:zero-shot}

To complement the pointmap probes, we use two zero-shot analyses to study how latent states and reconstructions evolve across multi-view transformer layers.

\paragraph{Representation similarity measure.}
To relate pointmap error evolution to changes in the latent state, we measure the calibrated linear Centered Kernel Alignment (CKA)~\cite{kornblith2019similarity,groger2026revisiting} between MVT block outputs. We compute the similarity between patch tokens extracted after the last residual connection of each block.

\paragraph{Evaluating emergent correspondences.}
We also evaluate the emergence of correspondences, a core cue in SfM pipelines, during reconstruction. Following~\cite{an2024zeroco}, we extract patch-level zero-shot correspondences from cross-view attention (CVA) maps. Let \((a, b)\) be a pair of views, and let \(p^a\) be a query patch in view \(a\) with a ground-truth correspondence in view \(b\) (``stereo patch''). For each CVA layer \(\ell\) and head \(h\), we predict its match as the target patch \(p^b\) with the maximum attention score:

\begin{equation}
    \hat{p}^b_{\ell,h} = \arg\max_{p^b} A^{a \rightarrow b}_{\ell,h}\left(p^a, p^b\right).
\end{equation}

For VGGT and DA3, we decompose each global attention map into within-view and cross-view maps, and use the cross-view map. For DUSt3R, we use the cross-attention maps directly and follow the heuristic from~\cite{an2024zeroco}, suppressing the first sequence token when it dominates the attention ranking. We report per-head correspondence accuracy over all stereo query patches and summarize each layer by its best-performing head. We obtain ground-truth patch correspondences by aggregating pixel correspondences derived from reprojected ground-truth 3D geometry.

The CKA and correspondence evaluation procedures are detailed in Appendix~\ref{subsec:app-cka-measure} and Appendix~\ref{subsec:app-corresp_prob}.

\subsection{Multi-View Transformer Ablations}\label{subsec:mvt_ablations}

To study how MVT design choices affect 3D state evolution and emergent correspondences, we implement multi-view transformers in a unified codebase that allows us to vary three components from~\cref{tab:model_comparison}: the encoder, cross-view attention mechanism, and weight sharing. We otherwise use a VGGT-style MVT with alternating single-view and cross-view attention. Following MUSt3R~\cite{must3r_cvpr25}, we omit RoPE from cross-view attention. We use DUSt3R's pointmap output parameterization. The architecture is detailed in Appendix~\ref{subsec:app-mvt-training}.
\begin{figure}[!t]
    \centering
    \begin{subfigure}[t]{0.49\textwidth}
        \centering
        \includegraphics[width=\linewidth]{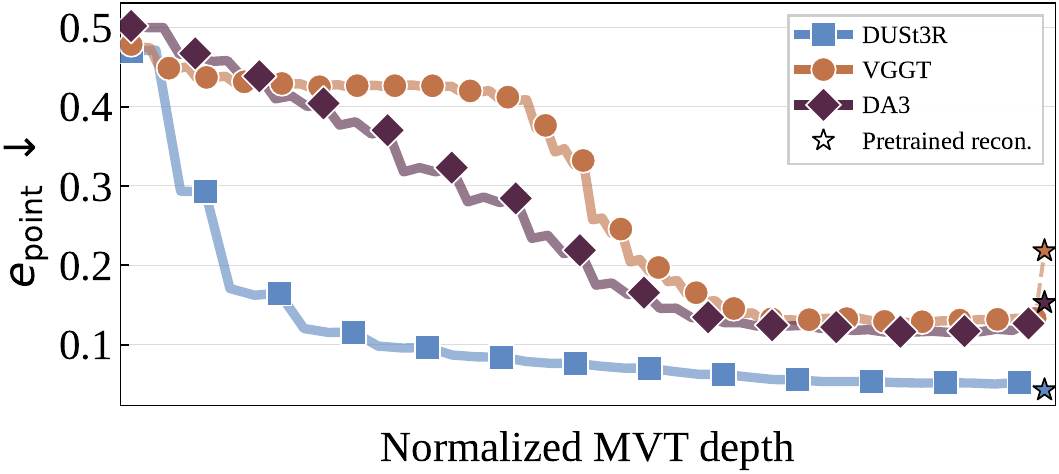}
        \caption{\textbf{Layer-wise reconstruction error.} \(e_{\mathrm{point}}\) evaluated after each residual update.}
        \label{fig:error_pointmap_main}
    \end{subfigure}\hfill
    \begin{subfigure}[t]{0.49\textwidth}
        \centering
        \includegraphics[width=\linewidth]{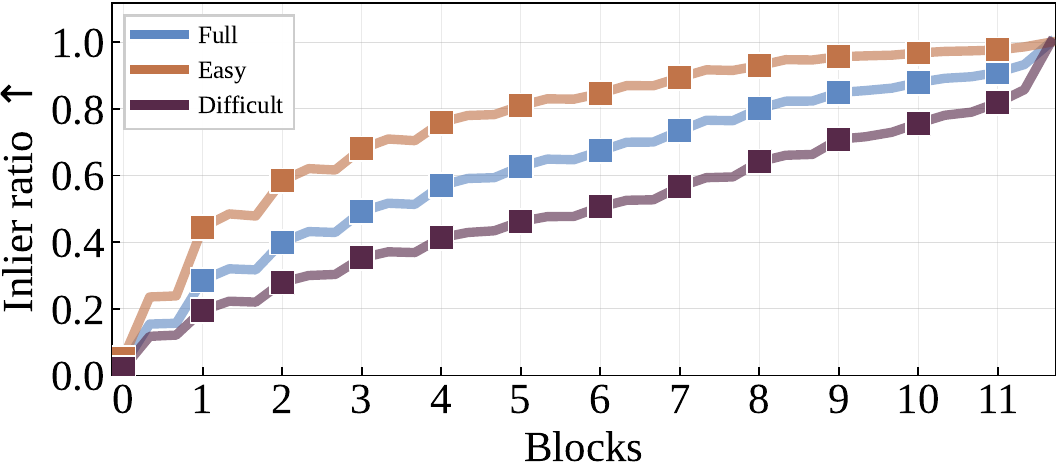}
       \caption{\textbf{Convergence speed vs.\ difficulty (DUSt3R).} Inlier ratio within \(5\%\) of the final probe reconstruction.}
        \label{fig:convergence_speed}
    \end{subfigure}
    
    \vspace{0.5em}
    
    \begin{subfigure}[t]{\textwidth}
        \centering
        \includegraphics[width=\linewidth]{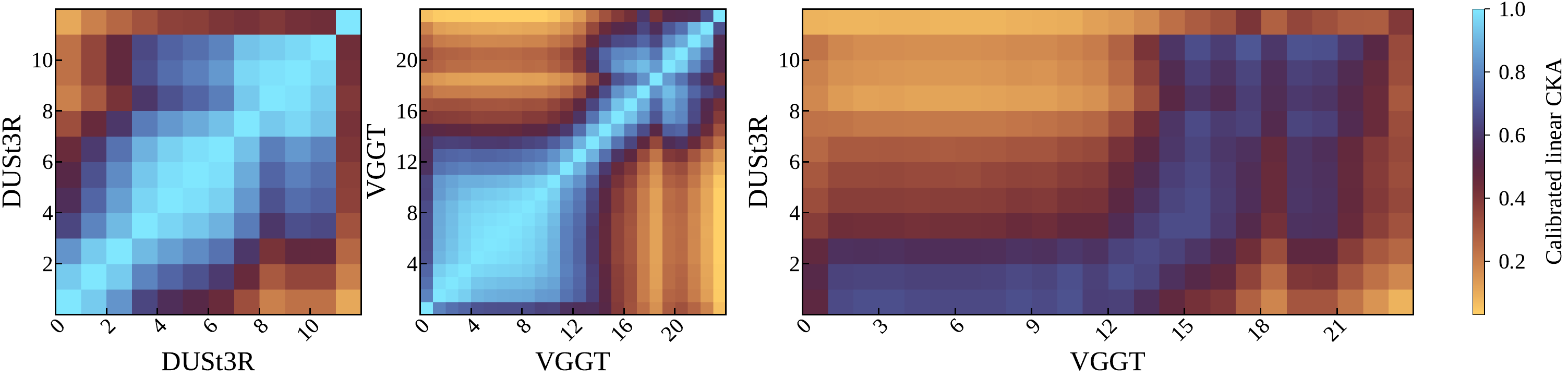}
        \caption{\textbf{Representation similarity.} Calibrated centered kernel alignment (CKA) across MVT block outputs.}
        \label{fig:cka_view2}
    \end{subfigure}
    \caption{\textbf{Second-view error and state evolution on HM3D.} \textbf{(a)} Pointmap test error relative to ground truth after each residual update; markers indicate block inputs. \textbf{(b)} Inlier ratio relative to the final probe reconstruction, grouped by sample difficulty. \textbf{(c)} Calibrated CKA similarity between MVT block outputs. These highlight the progressive nature of reconstruction, where refinement speed depends on difficulty and phase transitions in error align with internal representation dynamics.}    
    \label{fig:error_layers}
    \vspace{-0.5em}
\end{figure}

\section{Experiments}
Our experiments investigate three questions related to how MVTs address 3D reconstruction:
\begin{enumerate}
    \item How does the internal state evolve across the multi-view transformer?
    \item What is the role of individual layers?
    \item Does the network utilize correspondences?
\end{enumerate}

\paragraph{Datasets.}
We primarily train probes on 100,000 rendered images from the HM3D dataset~\cite{ramakrishnan2021hm3d} and evaluate the main metrics on 500 image pairs from held-out scenes. We additionally train and evaluate probes on BlendedMVS~\cite{yao2020blendedmvs}. For qualitative evaluation, we use real-world examples from RealEstate10K and CO3Dv2~\cite{re10k,reizenstein21co3d}, both within the models' training distribution. We additionally evaluate zero-shot correspondence matching on object-centric data, using GSO~\cite{downs2022google} and DTC~\cite{Dong_2025_CVPR} objects rendered against procedurally generated backgrounds in Blender; see appendix for details.

\paragraph{Training and implementation details.}
We parametrize each probe as a five-layer MLP with ReLU activations and hidden dimension 512, and train all probes for four epochs. We evaluate the official pretrained checkpoints of DUSt3R, VGGT, and Depth Anything 3 (DA3). Our default MVT implementation uses 12 alternating-attention blocks and a linear output head. The MVT is trained from scratch, while the pretrained encoder remains frozen. See Appendices \ref{subsec:app-models} and \ref{subsec:app-mvt-training} for details.

\subsection{How Does the Internal State Evolve Across Multi-View Transformer Layers?}\label{subsec:state}
\paragraph{Feed-forward 3D reconstruction with MVTs is internally progressive.}
We first analyze the evolution of the pointmap error \(e_{\mathrm{point}}\). On average, most MVT layers reduce the error (\cref{fig:error_pointmap_main}), consistent with prior analyses of residual networks and transformers~\cite{alain2018probing,jastrzębski2018residual,geva2022transformer,dar2023analyzing,molina2023traveling,belrose2023elicitinglatentpredictionstransformers}. This layer-wise improvement has a direct geometric interpretation: although multi-view transformers reconstruct 3D scenes in a single forward pass, they \textit{progressively refine} geometry through several internal updates rather than in a single step. This progression is analogous to iterative refinement but differs from iterative optimization approaches used in SfM~\cite{schoenberger2016sfm,pan2024glomap}, in that each layer applies a \textit{distinct} learned update rather than repeatedly applying a fixed rule to optimize an explicit objective.

\paragraph{Convergence speed depends on sample difficulty.}
\Cref{fig:vis_hard_easy} illustrates this progressive refinement. It shows that reconstructions of easy and difficult image pairs improve at different rates. At the output of the shared encoder, the network already provides a strong geometric estimate of the first view, suggesting that monocular depth and camera intrinsics are largely encoded before multi-view interaction. In the simple example, the rotation is resolved after the first MVT block, and later blocks align scale and refine translation. In the difficult example, the correct pose emerges only after several blocks, as second-view pointmaps gradually move toward the final estimate. This indicates that MVT depth acts as an iteration budget in optimization: additional layers let difficult samples keep refining when early updates are insufficient. Quantitatively, the DUSt3R inlier ratio with respect to the final probe estimate rises from 5 to nearly 50\% after the first MVT block on the easy subset. It remains less than half as high on the difficult subset and at least 20\% below the easy subset until the last block (\cref{fig:convergence_speed}). As observed in~\cite{lindenberger2023lightglue}, convergence speed depends on sample difficulty. Easy samples converge in one or a few blocks, whereas harder samples may require more blocks to converge.

\begin{figure}[!t]
    \centering
    \includegraphics[width=1\textwidth]{figures/main/low_res/iterative.pdf}
    \caption{\textbf{Pointmap visualizations for easy vs.\ hard samples (DUSt3R).} We visualize pointmap predictions across MVT blocks for an easy example (top) and a challenging one (bottom). The first view is shown in blue and the second in yellow. The yellow-tinted patches progressively move through 3D space toward their final positions in a semi-rigid manner. While early layers already yield a reasonable reconstruction for simple cases, harder samples require refinement across more layers.}
    \label{fig:vis_hard_easy}
\end{figure}

\paragraph{Error and state dynamics.}
We observe in \cref{fig:error_pointmap_main} that error profiles differ across models. DUSt3R's error drops sharply when multi-view information enters the MVT, DA3 improves more gradually, and VGGT remains nearly flat before an abrupt decrease. To relate these regimes to the evolution of the latent state, we analyze CKA similarities between blocks. For the pointmap-based transformers DUSt3R and VGGT, CKA similarity exhibits a phase structure~\cite{kornblith2019similarity,raghu2021vision,jacobs2026block} that mirrors the phase structure observed in the pointmap error curves (\cref{fig:error_pointmap_main} and \ref{fig:cka_view2}). This confirms that the probes track meaningful changes in the models' internal 3D state. This structure also appears in cross-model similarity between DUSt3R and VGGT: blocks within the same error regime, before or after the sharp error decrease, are more similar than blocks from different regimes. This suggests that the two models may follow similar reconstruction stages despite different convergence dynamics.

\paragraph{Explaining fast and slow dynamics.}
Output head design, more than MVT architecture, explains differences in error reduction speed across depth. To test this, we train MVTs from scratch on HM3D with either a final-layer or a multilayer head. Across architectures, final-layer heads yield rapid error reduction, whereas multilayer heads yield slower refinement (Appendix~\ref{subsec:app-output-ablation}, \cref{fig:app-head-training}). Models such as DUSt3R, trained first with a final-layer head and then with a multilayer head, retain their rapid initial reduction. This explains DUSt3R's faster convergence relative to VGGT and DA3, which are trained only with multilayer heads. We hypothesize that shortcuts to the output head redistribute reconstruction across MVT layers compared with standard residual networks~\cite{He_2016_CVPR,jastrzębski2018residual}.

\takeaway{Multi-view transformers progressively refine geometry internally, despite using a single forward pass. The final reconstruction emerges earlier for easy samples than for hard ones, as early as the first few layers in MVTs trained with final-layer output heads.}

\paragraph{Additional results.}
Appendix~\ref{sec:app-additional-res} shows that probing after residual updates and using nonlinear probes are essential for obtaining stable and interpretable layer-wise trajectories. It also reports detailed per-model error and convergence curves, CKA analyses, and output head ablations.

\subsection{What Is the Role of Individual Layers?}

To understand the role of individual layers, we inspect their action on probed intermediate pointmaps.

\begin{figure}[!t]
    \centering
    \includegraphics[width=1\textwidth]{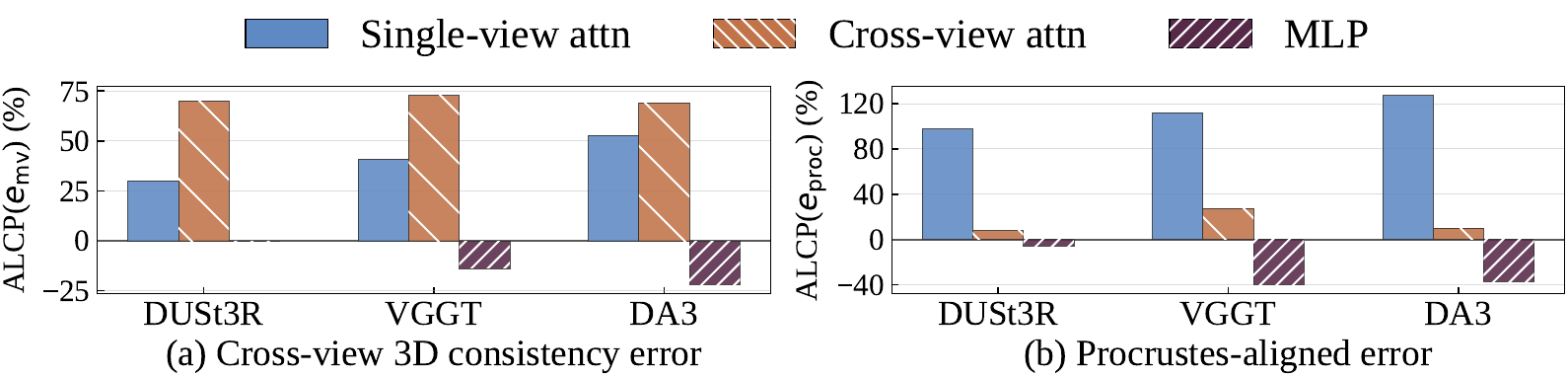}
    \caption{\textbf{Layer role analysis on HM3D.} Contribution of each layer type to second-view error reduction across the MVT for \textbf{(a)} \(e_{\mathrm{mv}}\) and \textbf{(b)} \(e_{\mathrm{proc}}\). Cross-view attention drives multi-view 3D consistency improvements across the MVT, while single-view attention refines non-reference geometry.}
    \label{fig:layer_role}
    \vspace{-1.5em}
\end{figure}

\paragraph{Cross-view attention improves multi-view consistency.}
\Cref{fig:vis_hard_easy} shows that the reference-view pointmap remains stable across MVT layers, while the non-reference view pointmap is progressively registered into the reference coordinate frame. Within this process, cross-view attention layers improve multi-view consistency by bringing corresponding point predictions closer together, as illustrated in appendix \cref{fig:layer-role-viz}. We quantify this effect with the accumulated layer contribution \(\text{ALCP}(e_{\mathrm{mv}}, \mathcal F)\), which measures how much each layer type \(\mathcal F\) contributes to reducing the multi-view 3D consistency error across the MVT. In \cref{fig:layer_role}a, cross-view attention accounts for most of this reduction on HM3D, contributing approximately 70\% of the decrease across the MVT for all models.

\paragraph{Single-view attention improves non-reference view geometry.}
After cross-view attention registers non-reference views to the reference frame, single-view attention layers improve their intra-view geometry. \Cref{fig:layer-role-viz} illustrates this behavior. We quantify this effect using the accumulated layer contribution for the Procrustes-aligned pointmap error, \(\text{ALCP}(e_{\mathrm{proc}}, \mathcal F)\), shown in \cref{fig:layer_role}b on HM3D probes. Across all models, single-view attention accounts for most of the net reduction across the multi-view transformer, contributing close to or above 100\% of the reduction because other layer types can partially increase this error.

\begin{table}[!h]
    \centering
    \small
    \caption{\textbf{MLP ablation on HM3D.} Pointmap error \(e_{\mathrm{point}}\) for the 3D reconstruction evaluated on held-out scenes. Parentheses report relative error changes from the VGGT baseline; negative values indicate improvement. Parameter counts and FLOPs exclude the frozen encoder and output heads.}
    \label{tab:app-mlp-ablation}
    \begin{tabular}{lrrrrr@{.}l@{\hspace{0.35em}}l}
        \toprule
        \textbf{Variant} & \textbf{Blocks} & \textbf{Width} & \textbf{Parameters} & \textbf{FLOPs} & \multicolumn{2}{c@{\hspace{0.35em}}}{\(e_{\mathrm{point}}\!\downarrow\)} & \\
        \midrule
        VGGT\(\dagger\) & 12 & 1024 & 302.4M & 334.0G & 0 & 1202 & \\
        VGGT\(\dagger\) -- MLP-free & 12 & 1024 & 100.8M & 124.6G & 0 & 1378 & {\scriptsize ($+14.6\%$)} \\
        VGGT\(\dagger\) -- MLP-free, scaled & 24 & 1216 & 285.6M & 342.6G & \textbf{0} & \textbf{1135} & {\scriptsize ($-5.6\%$)} \\
        \bottomrule
        \addlinespace[0.25em]
        \multicolumn{8}{@{}l}{\footnotesize $^\dagger$ Our VGGT-style re-implementation trained from scratch.} \\
    \end{tabular}\par
    \vspace{-9pt}
\end{table}

\paragraph{Role of MLP layers.}
Unlike cross-view and single-view attention, MLP layers show no clear geometric role in our probes: on HM3D, they increase the probed reconstruction error on average across all three models (\cref{fig:app-layer-role-hm3d}). We test their necessity by training VGGT-style MVTs from scratch on HM3D with and without MLP layers. Removing MLP layers degrades performance by only 14.6\% while reducing the trainable parameter count by a factor of three (see~\cref{tab:app-mlp-ablation}). A deeper and wider attention-only MVT with approximately the same parameter count as the baseline outperforms it by 5.6\%. It also performs better than pretrained DA3 and VGGT on HM3D, but worse than DUSt3R. These results suggest that MLP layers are not essential for effective 3D reconstruction in this setting, and that reallocating their parameters to attention layers can even improve performance.

\takeaway{MVTs reconstruct 3D scenes progressively by alternating cross-view alignment to the reference view with single-view shape refinement in other views. Our probes reveal no geometric role for MLPs; attention-only models outperform models with MLPs in small-scale experiments.}

\paragraph{Additional results.}
Appendix~\ref{subsec:app-role} provides full per-layer curves for all metrics and evaluation settings. It shows that the same layer-role trends persist for DUSt3R probes trained and evaluated on BlendedMVS and extend to a four-view VGGT setting. It includes a perturbation analysis that supports the identified role of cross-view attention, and details the MLP ablations and results.

\subsection{Do Multi-View Transformers Utilize Correspondences?}

After establishing that cross-view attention (CVA) aligns points across views, we examine whether image correspondences emerge alongside this alignment. We visualize individual attention heads and find, across all models, heads that consistently attend to candidate matching patches in the other view. These have also been observed in~\cite{bratulic2025undvggt}. We refer to these heads as ``correspondence heads'' and provide detailed DUSt3R visualizations in Appendix~\ref{app:ca}.

\paragraph{Correspondences turn from semantic to geometric.}
\begin{wrapfigure}{r}{0.48\textwidth}
    \vspace{-1.5em}
    \centering
    \includegraphics[width=\linewidth]{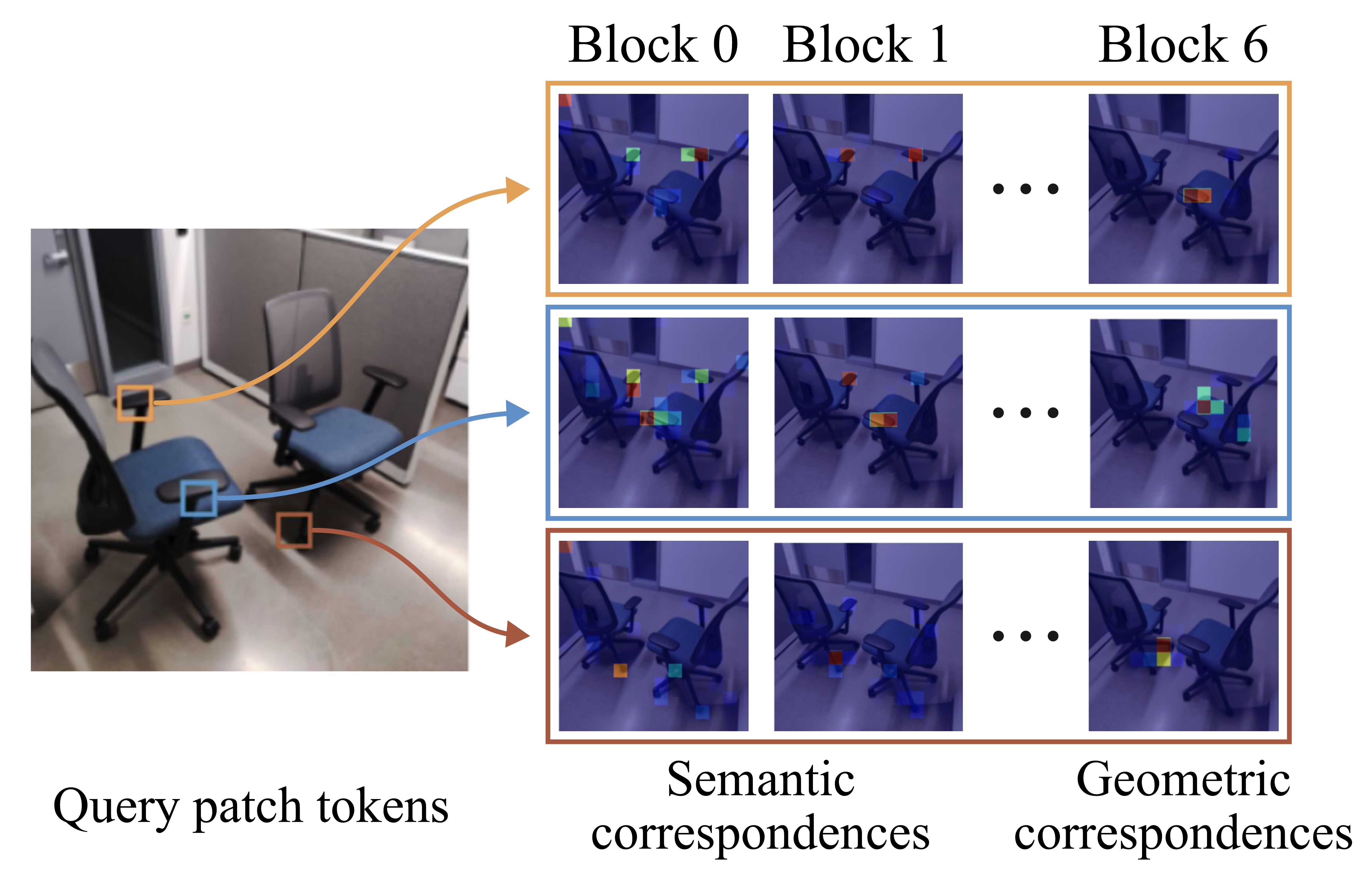}
    \caption{\textbf{Semantic-to-geometric refinement in DUSt3R.} Cross-view attention for three query patches shifts from semantically similar regions to geometrically corresponding patches.}
    \label{fig:corresp_head_main}
    \vspace{\dimexpr-1em-\baselineskip\relax}
\end{wrapfigure}
We first qualitatively examine correspondence heads using the example in \cref{fig:corresp_head_main}. In early DUSt3R MVT blocks, attention maps are consistent with \textit{semantic} correspondences~\cite{berg2005deformable}: they match patches with similar appearance or category. After several blocks, DUSt3R refines these matches into \textit{geometric} correspondences~\cite{elbanani2024probing}, attending to patches that observe the same 3D point.

\begin{figure}[!t]
    \centering
    \begin{minipage}[t]{0.49\linewidth}
        \vspace{0pt}
        \centering
        \includegraphics[width=\linewidth]{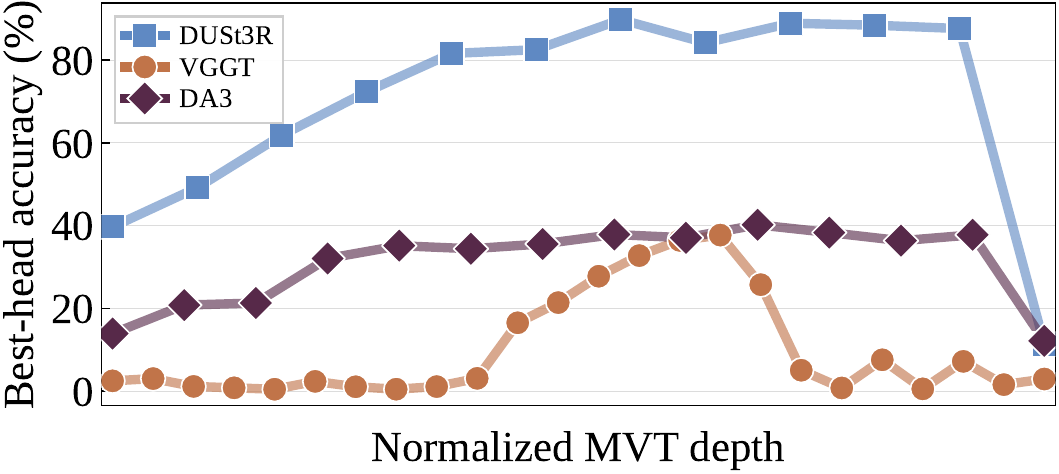}
        \caption{\textbf{Correspondence accuracy on HM3D.} Top-1 accuracy of the best-performing head in each second-view cross-view attention layer.}
        \label{fig:corresp_ref}
    \end{minipage}\hfill
    \begin{minipage}[t]{0.49\linewidth}
        \vspace{0pt}
        \centering
        \small
        \captionof{table}{\textbf{Cross-dataset correspondence accuracy.} Top-1 accuracy of the best-performing second-view cross-view attention head across scene- and object-centric datasets.}
        \label{tab:corresp_cross_dataset}
        \begin{tabular}{lccc}
            \toprule
            & \multicolumn{1}{c}{\textbf{Scene-level}} & \multicolumn{2}{c}{\textbf{Object-level}} \\
            \cmidrule(lr){2-2}\cmidrule(lr){3-4}
            \textbf{Model} & \textbf{HM3D} & \textbf{GSO} & \textbf{DTC} \\
            \midrule
            DUSt3R & 89.77\% & 70.31\% & 55.06\% \\
            VGGT   & 37.73\% & 59.33\% & 47.00\% \\
            DA3    & 40.22\% & 52.16\% & 42.77\% \\
            \bottomrule
        \end{tabular}
    \end{minipage}
\end{figure}

\paragraph{Measuring the refinement.} We quantify this refinement using the zero-shot correspondence estimator described in \cref{subsec:zero-shot}. \Cref{fig:corresp_ref} reports top-1 correspondence accuracy on HM3D for the best-performing head of each second-view CVA layer. All models improve through the first half of the MVT, most strongly DUSt3R: its accuracy rises from 40\% at the MVT input to nearly 90\% after six blocks. VGGT and DA3 start lower and reach nearly 40\%. As shown in~\cref{tab:corresp_cross_dataset}, correspondence accuracy in CVA layers also varies with scene type. DUSt3R's advantage over VGGT and DA3 narrows significantly on object-centric datasets: DUSt3R achieves \(2.4\times\) VGGT's accuracy on HM3D, compared with only \(1.2\times\) on object-centric datasets.

\paragraph{Correspondence ablation.} To understand the performance gap between DA3/VGGT and DUSt3R, we study how the pretrained encoder and cross-view attention mechanism affect correspondence accuracy. We train MVTs from scratch following the setup described in~\cref{subsec:mvt_ablations}. For scene-level evaluation, we train on HM3D and evaluate on held-out scenes. For object-centric evaluation, we train on DTC and evaluate on held-out objects, as well as on GSO objects. The results in Appendix~\cref{tab:corresp_mvt_ablation} show that CroCoV2-based models substantially outperform their DINOv2 counterparts, and cross-attention outperforms global attention in all but one setting. Overall, the model most similar to DUSt3R achieves the highest correspondence accuracy across all datasets, whereas the model most similar to VGGT and DA3 achieves significantly lower accuracy.

\takeaway{Emergent correspondences are refined across MVT cross-view attention layers. Encoder choice significantly affects this behavior, with CroCoV2 promoting strong correspondences.}

\paragraph{Additional results.} Appendix~\ref{subsec:app-corresp-result} provides correspondence accuracy for all views and heads, as well as visualizations of DUSt3R attention maps. We also include preliminary experiments replacing the attention maps of correspondence heads with ground-truth correspondences.

\section{Discussion}

\paragraph{Iterative refinement analogy.} Our results show that multi-view transformers resemble iterative SfM methods in several respects: geometry progressively improves, network depth acts as a refinement budget, and correspondences emerge and are used during reconstruction. However, MVT updates are computed by separate layers with \textit{distinct} parameters, unlike classical iterative optimization methods that repeatedly apply the \textit{same} update rule. Despite the similarities highlighted in our work, we therefore do not claim that an MVT's forward pass constitutes an iterative optimization procedure. Investigating the extent to which MVT updates can be understood as iterative refinement is a promising direction for future work, particularly in light of recent MVTs that achieve competitive reconstruction performance by reusing the same set of layers across multiple iterations~\cite{burzio2026dejaview}.

\paragraph{Implications for future model design.} While this work focuses on analysis rather than new model design, our results suggest several concrete directions for improving multi-view transformers. First, our small-scale experiments suggest that MLP layers may not be necessary and that deeper and wider attention-only MVTs can even improve performance. Second, our iterative view of MVTs is closely related to recent parameter-efficient approaches that repeatedly apply a fixed set of layers~\cite{burzio2026dejaview}. Finally, finding ways to use emergent, interpretable behaviors such as correspondences to enhance generic MVTs with explicit inductive biases after training is an interesting direction.

\section{Conclusion}
We introduced a layer-wise pointmap probing framework for studying how multi-view transformers encode and update 3D geometry during feed-forward reconstruction. Applying this framework to DUSt3R, VGGT, and Depth Anything 3, we found that decoded intermediate reconstructions improve over several layers rather than appearing only at the output. This progression varies across architectures and sample difficulty, suggesting that transformer depth acts partly as a learned refinement budget. Our analysis further shows that this refinement is structured. Cross-view attention primarily improves multi-view consistency by aligning stereo points, while single-view attention improves intra-view geometry in non-reference views. We also find that a subset of cross-view attention heads behaves like correspondence matchers, with correspondences becoming more geometric as reconstruction improves. Together, these results show similarities between multi-view transformers and traditional SfM pipelines. They provide a direct view of reconstruction dynamics inside multi-view transformers and suggest that pointmap probing can serve as a practical tool for analyzing future feed-forward 3D and 4D reconstruction models.

\paragraph{Acknowledgments} The authors are grateful to Lahav Lipson and Duc-Hai Pham for helpful discussions and for reviewing an earlier version of the manuscript.

VS was supported by the National Science Foundation under Grant No. 2211259, by the Singapore DSTA under DST00OECI20300823 (New Representations for Vision, 3D Self-Supervised Learning for Label-Efficient Vision), by the Intelligence Advanced Research Projects Activity (IARPA) via Department of Interior/ Interior Business Center (DOI/IBC) under 140D0423C0075, by the Amazon Science Hub, and by the MIT-Google Program for Computing Innovation. AT acknowledges the NVIDIA Academic Grant Program. MO was supported by the ERC Consolidator Grant 101087347 (VEGA), as well as gifts from Ansys Inc., and Adobe Research. This work was granted access to the HPC resources of IDRIS under the allocation 2025-AD011017075 made by GENCI.

\newpage
\printbibliography


\clearpage
\appendix

\section*{Appendix}

\subsection*{Overview of the Appendix}
This appendix provides implementation and evaluation details, additional experiments, and qualitative examples supporting the analyses in the main paper.

\paragraph{Section~\ref{sec:app-training-evaluation}: Training and evaluation details.}
We specify the datasets, probe architecture, training procedure, and pointmap metrics (\cref{subsec:app-datasets,app:probes,app:train_details,subsec:app-pointmap_eval_details}). We detail zero-shot representation similarity measure and correspondence evaluation (\cref{subsec:app-corresp_prob,subsec:app-cka-measure}). We finally describe the shared MVT architecture and training setup used in our design ablations (\cref{subsec:app-mvt-training}).

\paragraph{Section~\ref{sec:app-additional-res}: Additional results.}
We first examine probe capacity and probing granularity (\cref{app:ablation}), then analyze internal-state evolution, convergence by sample difficulty, and the effect of output head training (\cref{subsec:app-state}). We provide further quantitative and qualitative evidence for individual layer roles, including cross-dataset and four-view evaluations and the MLP ablation (\cref{subsec:app-role}). Finally, we report correspondence analyses and ablations (\cref{subsec:app-corresp-result}), followed by additional pointmap and attention-map visualizations (\cref{subsec:app-visualizations}).

\section{Training and Evaluation Details}\label{sec:app-training-evaluation}

\subsection{Models Studied}\label{subsec:app-models}

We probe models using their official checkpoints, available on HuggingFace:
\begin{itemize}
    \item DUSt3R: \texttt{naver/DUSt3R\_ViTLarge\_BaseDecoder\_512\_dpt}
    \item VGGT: \texttt{facebook/VGGT-1B}
    \item Depth Anything 3 (ViT-G): \texttt{depth-anything/DA3-GIANT-1.1}
\end{itemize}

\subsection{Datasets}\label{subsec:app-datasets}
\begin{figure}[!h]
    \centering
    \includegraphics[width=\linewidth]{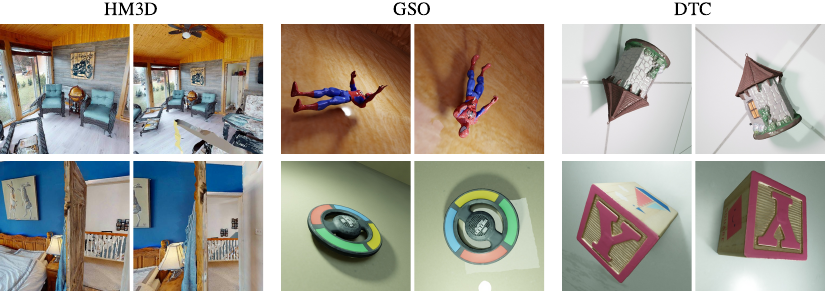}
    \caption{\textbf{Example image pairs.} Test pairs from the scene-level HM3D dataset and the object-centric GSO and DTC datasets.}
    \label{fig:app-dataset-pairs}
\end{figure}

\Cref{fig:app-dataset-pairs} shows example pairs from each dataset.

\paragraph{HM3D.} We train probes and evaluate the main metrics on image pairs rendered from HM3D scenes~\cite{ramakrishnan2021hm3d}, using held-out scenes for evaluation.

\paragraph{Object-centric datasets.} We render GSO~\cite{downs2022google} and DTC~\cite{Dong_2025_CVPR} objects with the same BlenderProc~\cite{Denninger2023} pipeline. Each scene contains a single object placed in a room, rescaled to a bounding-box diagonal of 2.7 scene units. We randomize the textures of the room surfaces, as well as the specularity, roughness, and metallic parameters of all materials. We place an emissive ceiling of random color and intensity and a randomly placed point light. For each object, we render two sequences of five \(720\times540\) views with a fixed focal length of 500 pixels. Cameras lie on a sphere of radius 3 scene units around the object, at elevations between \(30^\circ\) and \(89^\circ\), and point toward the object with a random roll. We split GSO by object into 540/60/151 train/val/test objects, and DTC by category into 1353/143/360 train/val/test objects.

\subsection{Probing Mechanism}\label{app:probes}

Our probing mechanism is illustrated in \cref{fig:probe}. Each probe is parameterized as a 5-layer MLP with ReLU activations and input LayerNorm~\cite{ba2016layer}.

\subsection{Training Details}\label{app:train_details}

\paragraph{Training loss.}

The probes are supervised via a confidence-weighted 3D regression loss introduced in DUSt3R \cite{dust3r_cvpr24}. For a given view, let \(\bar{X}\) denote the ground-truth pointmap provided over a subset of valid pixels \[
D \subseteq \{1,\ldots,W\} \times \{1,\ldots,H\}.
\]
We first express the ground-truth pointmaps of both views in the first-view coordinate system, and normalize the predicted and ground-truth pointmaps by their respective scale factors, each computed jointly over the valid pixels of both views:
\[
z = \operatorname{norm}\big(X^{(1)}, X^{(2)}\big) \quad \text{and} \quad \bar{z} = \operatorname{norm}\big(\bar{X}^{(1)}, \bar{X}^{(2)}\big),
\]
where \(D^{(v)}\) denotes the valid pixels of view \(v\) and
\[
\operatorname{norm}\big(X^{(1)}, X^{(2)}\big) = \frac{1}{|D^{(1)}| + |D^{(2)}|} \sum_{v \in \{1,2\}} \sum_{(i,j) \in D^{(v)}} \big\|X^{(v)}_{i,j}\big\|_2.
\]

The per-pixel regression loss for view \(v\) is then defined as
\begin{equation}
\ell_{\text{regr}}^{(v)}(i,j)= \left\|\frac{X_{i,j}^{(v)}}{z} - \frac{\bar{X}_{i,j}^{(v)}}{\bar{z}}\right\|_2.
\end{equation}
The final loss, computed over both views expressed in first-view coordinate system, is given by:

\begin{equation} \label{eq:confidence_loss}
L = \sum_{v \in \{1,2\}} \frac{1}{|D^{(v)}|} \sum_{(i,j)\in D^{(v)}} \left[ C^{(v)}_{i,j} \, \ell_{\text{regr}}^{(v)}(i,j) - \alpha \log\left(C^{(v)}_{i,j}\right) \right],
\end{equation}

where \(\alpha\) is a weighting hyperparameter that regularizes the confidence, and \(D^{(v)}\) denotes the valid pixels in view \(v\). We use \(\alpha=0.2\) in our experiments. Confidence scores are learned without explicit supervision. Each pixel's confidence value \(C^{(v)}_{i,j}\) modulates the loss to reduce the impact of unreliable or ambiguous predictions (e.g., sky or translucent regions).

\paragraph{Optimizer.} We use the AdamW optimizer \cite{loshchilov2018decoupled} with a learning rate of 5e-4 and a weight decay of 0.01.

\paragraph{Training details.} We train our probes for 4 epochs on each dataset on four Nvidia H100 GPUs.

\subsection{Pointmap Evaluation}\label{subsec:app-pointmap_eval_details}

\paragraph{Relative pointmap error.}
We measure pointmap accuracy with an absolute relative 3D error. Let \(X_n^{(v)}, \bar{X}_n^{(v)} \in \mathbb{R}^{H \times W \times 3}\) denote the predicted and ground-truth pointmaps for sample \(n\) and view \(v \in \{1,2\}\), expressed in the first-view coordinate frame, and let \(D_n^{(v)}\) denote the valid pixels. Following the MapAnything evaluation protocol, we first normalize predicted and ground-truth pointmaps by their average valid 3D distance. The per-pixel pointmap error for pixel coordinates \((i,~j)\) is
\begin{equation}\label{eq:app-pointmap-absrel}
    r_{n,v}(i,j) =
    \frac{\left\lVert X_n^{(v)}(i,j) - \bar{X}_n^{(v)}(i,j) \right\rVert_{2}}
    {\left\lVert \bar{X}_n^{(v)}(i,j) \right\rVert_{2}}.
\end{equation}
We first average over valid pixels within each sample and view,
\begin{equation}
    e_{\mathrm{point}}(n,v) =
    \frac{1}{|D_n^{(v)}|}\sum_{(i,j)\in D_n^{(v)}} r_{n,v}(i,j),
\end{equation}
and then average these per-sample errors over the dataset. When reporting a global two-view error, we average the two view errors for each sample before averaging over samples.

\paragraph{Procrustes alignment.}\label{app:procrustes-align}
To evaluate pointmaps up to a rigid transformation ambiguity, we align predicted and ground-truth pointmaps \(X, \bar X\) using a Procrustes alignment~\cite{schonemann1966generalized}, following~\cite{sorkine2017least,smith24flowmap}.
This corresponds to solving the following weighted Procrustes problem:
\begin{equation}
    \arg\min_{P \in SE(3)} \;\bigl\|\,W^{1/2}\bigl(X - P\,\bar{X}\bigr)\bigr\|_F^2,
\end{equation}
where \(W\) is a diagonal weight matrix. We use uniform weights, as we do not observe a significant improvement when using the confidence weights predicted by our probes. We report the resulting aligned error as \(e_{\mathrm{proc}}\).

\paragraph{Multi-view 3D consistency error.} To measure consistency between predictions for the same 3D point across views, we warp each view's predictions onto the other with the pixel correspondences from~\cref{subsec:app-corresp_prob}. We then compute the pointmap error \(e_{\mathrm{point}}\) (\cref{eq:app-pointmap-absrel}) using the warped predictions as ground truth. To measure the model's ability to estimate consistent reconstructions within a shared coordinate frame, we do not apply scale normalization before computing the relative error \(e_{\mathrm{point}}\).

\paragraph{Difficulty subsets.}
We define sample difficulty from the final reconstruction error of each pretrained model. For each test image pair, we first evaluate the final pretrained reconstruction and compute the per-sample pointmap error \(e_{\mathrm{point}}\) against ground truth. Since evaluation uses symmetrized image pairs, we compute this error in both directions and average the two values to obtain one scalar difficulty score per sample. We then rank test samples by this score and define the easy and hard subsets as the bottom and top 20\% of the error distribution, respectively. On our 500-pair test set, each subset contains 100 samples.

\subsection{Measuring Representation Similarity.}\label{subsec:app-cka-measure}
We measure the linear Centered Kernel Alignment (CKA)~\cite{kornblith2019similarity} by treating each patch token in the dataset as an individual sample. Instead of reporting a debiased per-batch CKA estimate as in previous work~\cite{nguyen2021do}, we compute the exact CKA over the entire dataset via online accumulation across batches. We report the CKA after the calibration procedure described in~\cite{groger2026revisiting} to remove the confounding effect of feature dimensionality, which can induce spurious correlations.

To do so, we first compute the uncalibrated CKA score \(s\) by accumulation over the entire dataset. During this process, for every pair of MVT layers, we randomly sample and store 500 patch tokens for calibration. For a given pair, we then use these samples to compute a calibration threshold \(\tau_{\alpha=0.5}\), and compute the calibrated score following their procedure as:
\begin{equation*}
    s_{\text{cal}} = \max\left(\frac{s - \tau_{\alpha=0.5}}{1 - \tau_{\alpha=0.5}}, ~0 \right)
\end{equation*}

\subsection{Zero-shot Correspondence Extraction and Evaluation}\label{subsec:app-corresp_prob}

Following~\cite{an2024zeroco}, we evaluate whether MVT cross-view attention maps can serve as zero-shot correspondence estimators. For each cross-view attention layer \(\ell\), attention head \(h\), and direction \(a \rightarrow b\), we denote the corresponding attention map by \(A^{a \rightarrow b}_{\ell,h}\). For a source patch \(p^a\), the top-1 predicted match is the target patch with maximum attention:
\begin{equation*}
    \hat{p}^b_{\ell,h} = \arg\max_{p^b} A^{a \rightarrow b}_{\ell,h}(p^a, p^b).
\end{equation*}

\paragraph{Ground-truth patch correspondences.}
Ground-truth pointmaps are projected across views using the camera poses and intrinsics. We then compute symmetric mutual correspondence masks with a tolerance of \(\tau=5\) pixels, since we find smaller thresholds to be too aggressive on HM3D, possibly due to rendering artifacts. We convert these pixel-level correspondences to patch-level correspondences by voting. For each source patch, all pixels in that patch vote for the target patch containing their projected correspondence. Any target patch with at least three votes is considered a valid match. This one-to-many mapping reflects the ambiguity of the patch-level correspondence matching task, with several patches forming potentially valid matches.

\paragraph{Correspondence evaluation.}
For each valid source patch, layer, head, and direction, we measure top-\(k\) matching accuracy: a prediction is correct if one of the ground-truth target patches appears among the \(k\) most-attended target patches. We favor this metric over distance threshold metrics, since we evaluate coarse correspondences at the patch level. We evaluate \(k \in \{1,3,5\}\) and aggregate counts over all valid query patches in the evaluation set. For DUSt3R, we apply the first-token heuristic of~\cite{an2024zeroco}: if the first target token is the most-attended token for a query, its attention score is set to zero before computing the top-\(k\) ranking. For VGGT and DA3, we first split the global attention map into within-view and cross-view blocks, and evaluate only the cross-view block. We report accuracies separately for both directions, yielding one accuracy score per layer and head. We summarize each layer by reporting its best-performing head.

\subsection{MVT Design and Ablations}\label{subsec:app-mvt-training}

\begin{figure}[!t]
    \centering
    \begin{minipage}[t]{0.49\textwidth}
        \centering
        \includegraphics[width=\linewidth]{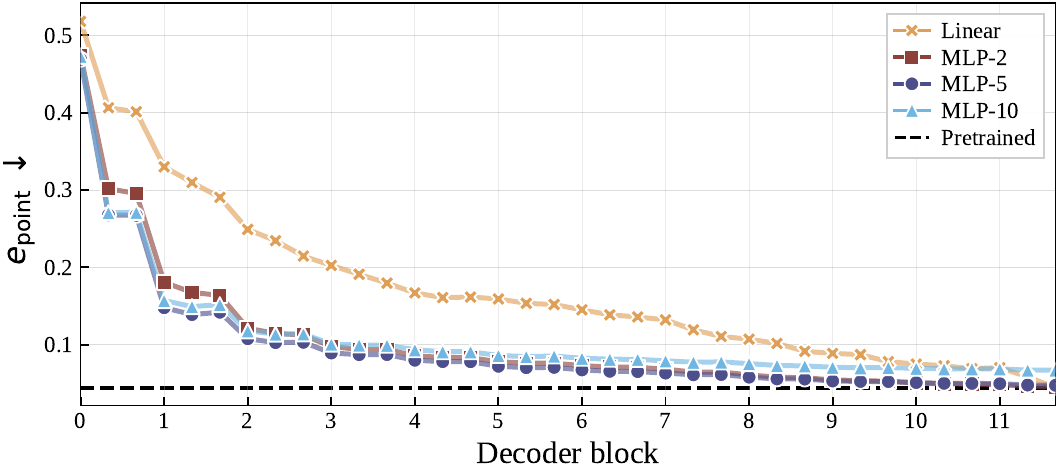}
        \caption{\textbf{Ablation study on probing head strength.} 
        We measure \(e_{\mathrm{point}}\) at the outputs of the MVT blocks for the second view, comparing different probing head architectures and capacities.}
        \label{fig:strength}
    \end{minipage}\hfill
    \begin{minipage}[t]{0.49\textwidth}
        \centering
        \includegraphics[width=\linewidth]{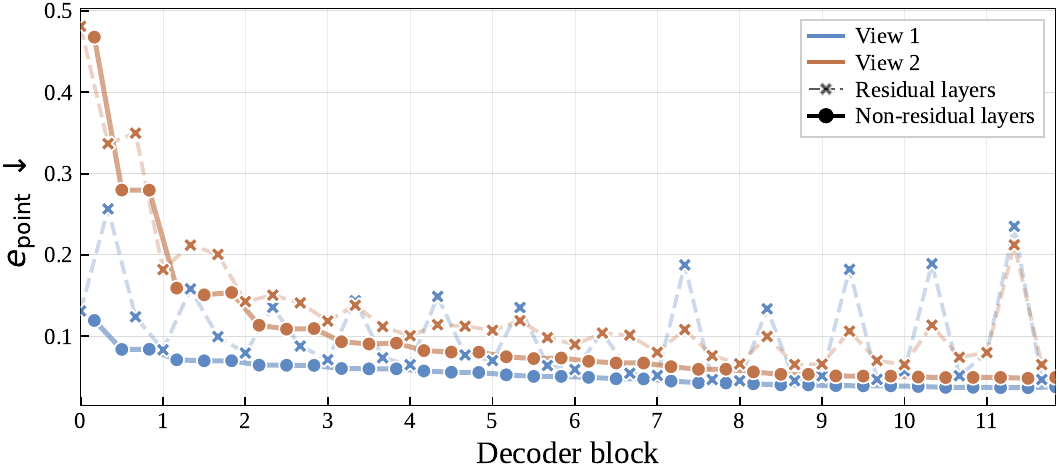}
        \caption{\textbf{Probing granularity ablation.} 
        We measure \(e_{\mathrm{point}}\) on HM3D. We plot the error at the output of residual layers (dashed) and after skip connections (solid). The error decreases almost monotonically after skip connections, while other operations exhibit more irregular variations.}
        \label{fig:skipvsnorm}
    \end{minipage}
\end{figure}

\paragraph{Shared architecture.}
We implement the MVT design space ablation described in~\cref{subsec:mvt_ablations} in a unified codebase, varying the main components listed in~\cref{tab:model_comparison}: pretrained encoder, cross-view attention mechanism, and view-wise weight sharing. The standard configuration closely follows VGGT but uses a smaller architecture. It uses 12 alternating-attention blocks of width 1024 with 16 attention heads. Each block contains single-view attention followed by cross-view attention, with each attention operation followed by a two-layer MLP with GELU activations~\cite{hendrycks2016gaussian}. We use pre-LayerNorm~\cite{ba2016layer}, query/key normalization~\cite{henry2020query}, and LayerScale~\cite{touvron2021going} initialized to 0.01. Rotary position embeddings~\cite{su2024roformer} are applied to single-view attention but omitted from cross-view attention, following MUSt3R~\cite{must3r_cvpr25}. Our implementation naturally supports any number of views.

\paragraph{Encoders and registers.}
We use pretrained CroCoV2 or DINOv2 ViT-L encoders and keep their weights frozen. Four learned register tokens~\cite{darcet2024vision} are concatenated to each view's token sequence, with separate parameters for the reference and non-reference views even when MVT weights are shared. As noted in VGGT, these distinct register tokens act as a reference view encoding and enable the model to identify the reference view even when MVT and output head weights are shared.

\paragraph{Output head and 3D parameterization.}
Unless otherwise specified, we use a linear output head applied to the final block output. We also implement a dense multilayer convolutional head adapted from VGGT-Omega~\cite{wang2026vggtomega}. It aggregates block outputs at relative depths \(0.25, 0.5, 0.75, 1.0\), rather than using only the final state. By default, we use DUSt3R's 3D output parameterization and predict pointmaps in the first-view coordinate frame and their associated confidence maps. We also implement VGGT's output parameterization with pointmaps, depth, and poses.

\paragraph{Training details.}
We train at a resolution of \(224\times224\), using a batch size of 16. We use DUSt3R's confidence-weighted pointmap regression loss described in~\cref{eq:confidence_loss}. We also implement VGGT's loss and support training with an arbitrary number of views. The MVT and output head are randomly initialized and trained with the AdamW~\cite{loshchilov2018decoupled} optimizer, using a base learning rate of \(10^{-4}\), weight decay of 0.01, and gradient-norm clipping at 1. We train for 10 epochs, with half an epoch of linear warmup followed by cosine learning-rate decay to \(10^{-6}\).

\section{Additional Results}\label{sec:app-additional-res}

\subsection{Ablations}\label{app:ablation}

 \begin{figure}[!t]
        \centering
        \includegraphics[width=0.6\linewidth]{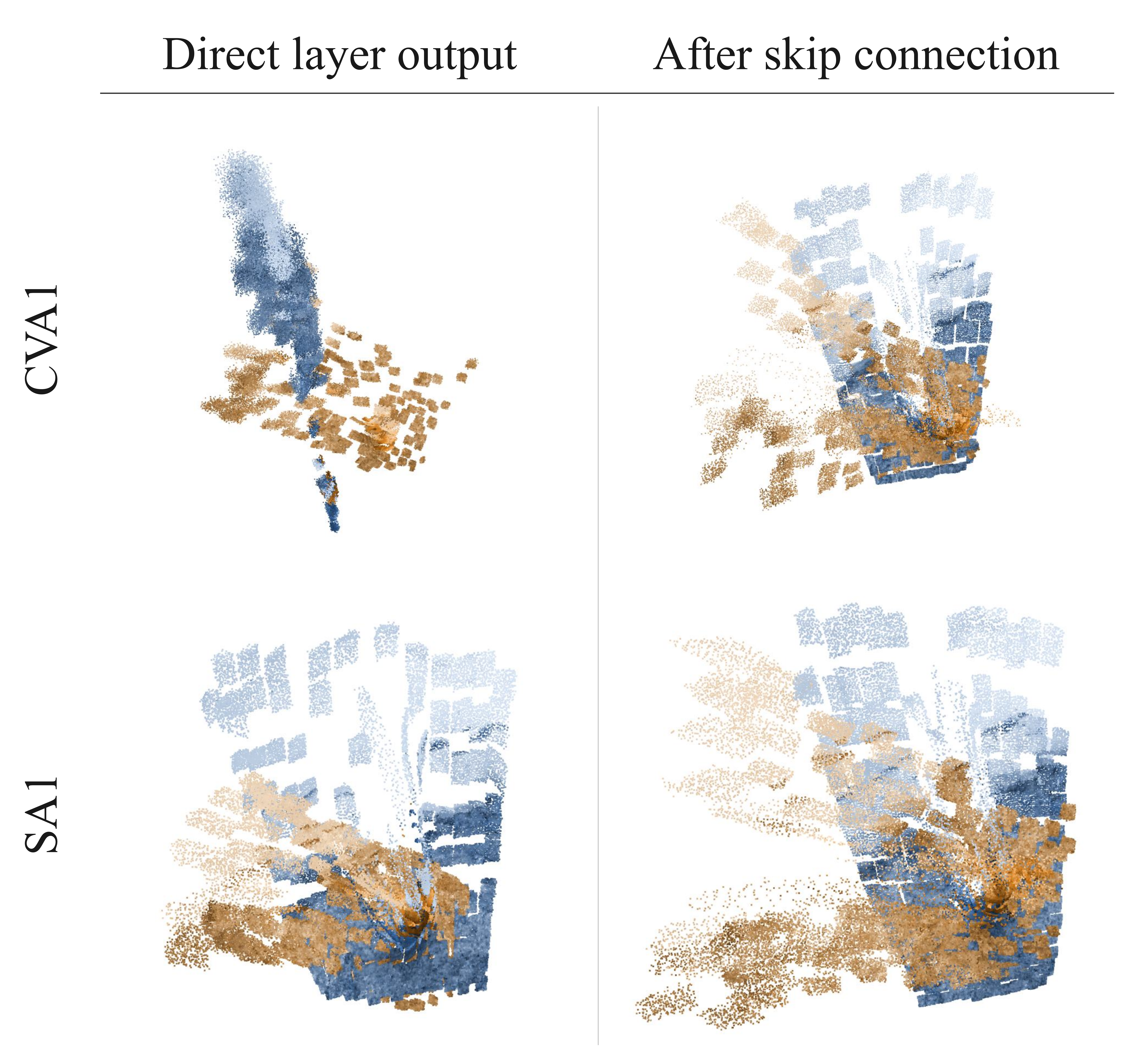}
        \caption{\textbf{Visual comparison of pointmaps} probed after skip connections vs direct layer outputs. We observe that probing after skip connections yields clearer and more interpretable geometry.}
        \label{fig:skipvsnorm_vis}
\end{figure}

\paragraph{Probe capacity.}\label{subsec:app-abla-cap}
We find that using non-linear probes yields considerably cleaner and more interpretable pointmaps compared to a linear head. This also results in a faster decrease of \(e_{\mathrm{point}}\) over the blocks, as reported in \Cref{fig:strength}. Furthermore, we find that using more than 5 layers does not affect the results.

\paragraph{Probing granularity.}
We verify the hypothesis of skip connections as a state carrier by probing the direct outputs of every individual layer before skip connections. \Cref{fig:skipvsnorm} shows the comparison of \(e_{\mathrm{point}}\) when probed from the skip connection versus when probed directly at the output of the layer. The error after skip connections is clearly more stable. Our visualization in \Cref{fig:skipvsnorm_vis} also shows that the direct outputs are less interpretable by pointmap probing. This indicates that the individual layers alter the state in a way that does not necessarily have clear geometric meaning, while the state update from the skip connection does. 

\subsection{Evolution of the internal state}\label{subsec:app-state}

\paragraph{Full pointmap error curves.}
In \cref{subsec:app-role}, we provide the full layer-wise pointmap curves used to analyze individual layer roles. We report the HM3D results used in the main text, DUSt3R probes trained and evaluated on BlendedMVS, HM3D-trained probes evaluated zero-shot on the MapAnything benchmark datasets (TartanAirv2~\cite{wang2020tartanair}, ScanNet++~\cite{yeshwanthliu2023scannetpp}, and ETH3D~\cite{schoeps2017cvpr}), and a four-view VGGT setting trained and evaluated on HM3D. These additional settings test whether the same layer-wise trends persist across datasets and beyond the two-view setup.

\begin{figure}[p]
    \centering
    \includegraphics[width=0.82\linewidth]{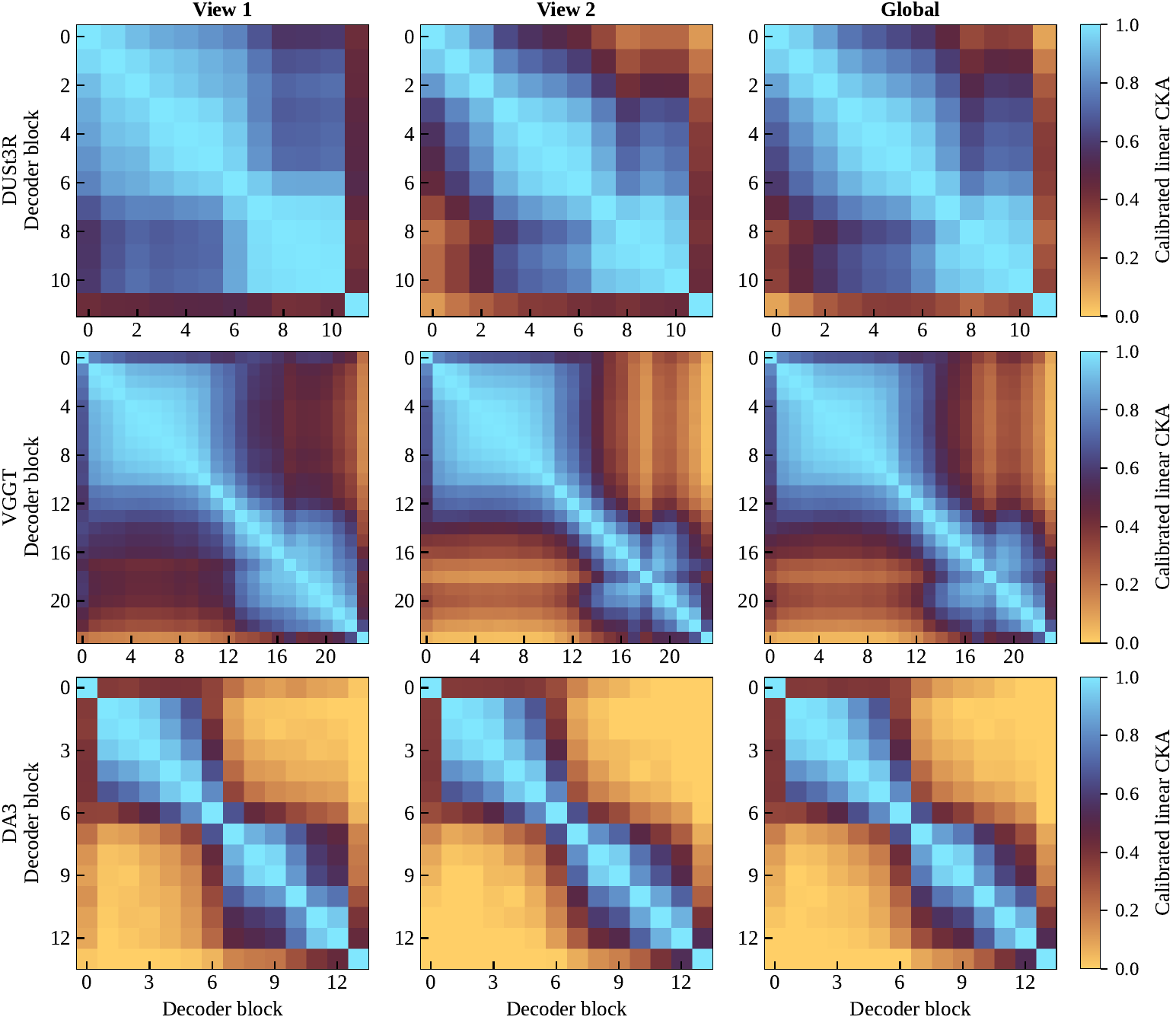}
    \caption{\textbf{Centered Kernel Alignment (CKA) between MVT blocks.}
    Pairwise calibrated CKA scores between the last layer of each MVT block.
    Results are computed over the validation set and reported separately for features from view~1 and view~2, as well as jointly over both views (``Global''), where features from both views are concatenated prior to CKA computation.}
    \label{fig:cka}
    \bigskip

    \centering
    \includegraphics[width=0.82\linewidth]{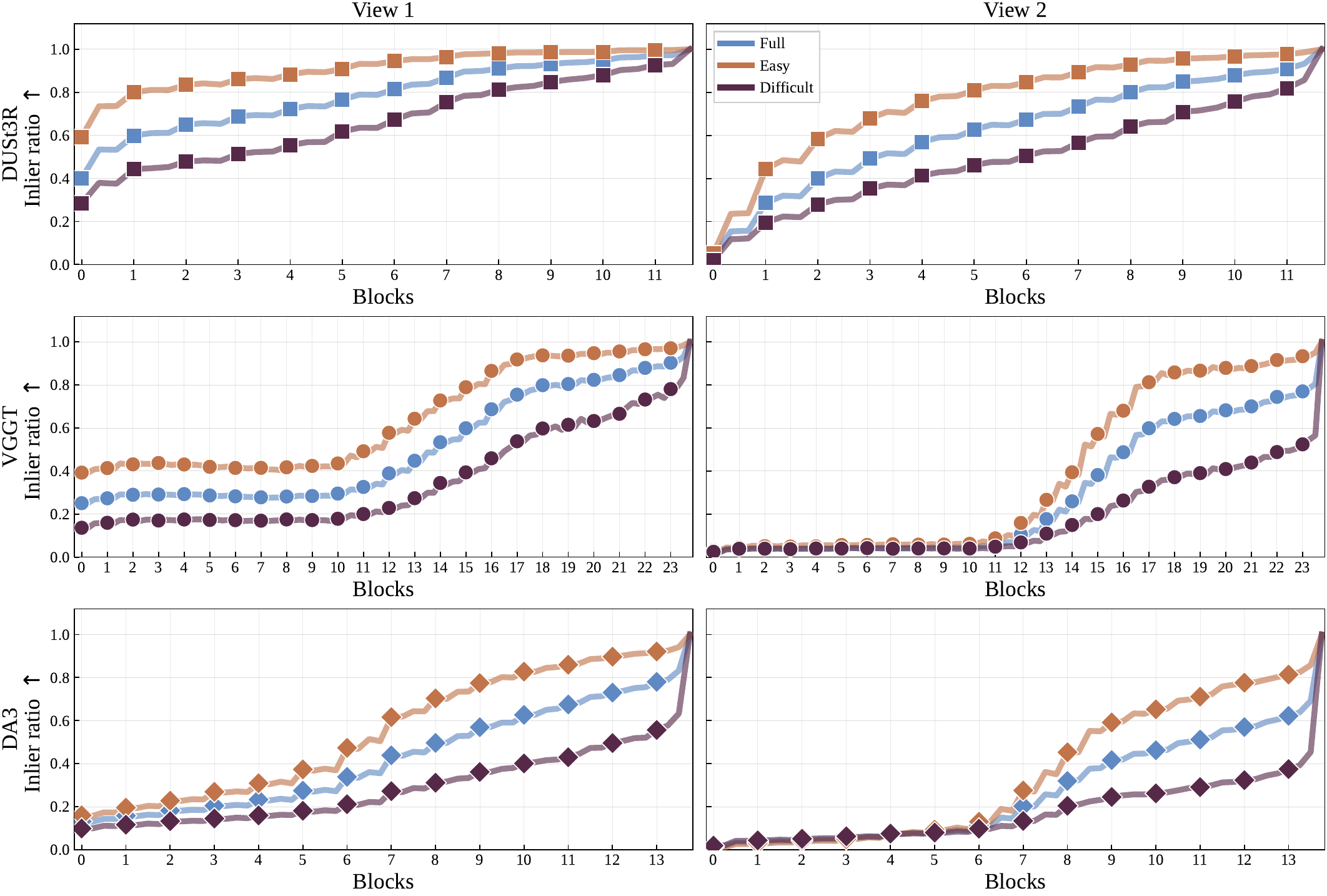}
    \caption{\textbf{Convergence by sample difficulty.} Inlier ratio within \(5\%\) relative error of the final probe reconstruction, averaged over the full dataset or each model's easiest and hardest \(20\%\) of test samples.}
    \label{fig:app-convergence-difficulty}
\end{figure}

\paragraph{Centered kernel alignment measure.} In \Cref{fig:cka}, we report Centered Kernel Alignment (CKA) scores between MVT blocks for several multi-view transformers, computed over the validation set after the calibration procedure described in \cref{subsec:app-cka-measure}. As computing similarities between all pairs of layers is expensive, we visualize similarities at the block-level by measuring similarity between the last layer of each block.

\paragraph{Convergence by sample difficulty.}
\Cref{fig:app-convergence-difficulty} compares convergence across blocks toward the final probe reconstruction for DUSt3R, VGGT, and DA3, using all samples and the easiest and hardest \(20\%\) defined in~\cref{subsec:app-pointmap_eval_details}. Across models and views, easy samples converge faster on average. VGGT and DA3 do not converge before the final layer at the \(5\%\) error threshold, with second-view inlier ratios below \(60\%\).

\begin{figure}[!t]
    \centering
    \includegraphics[width=\linewidth]{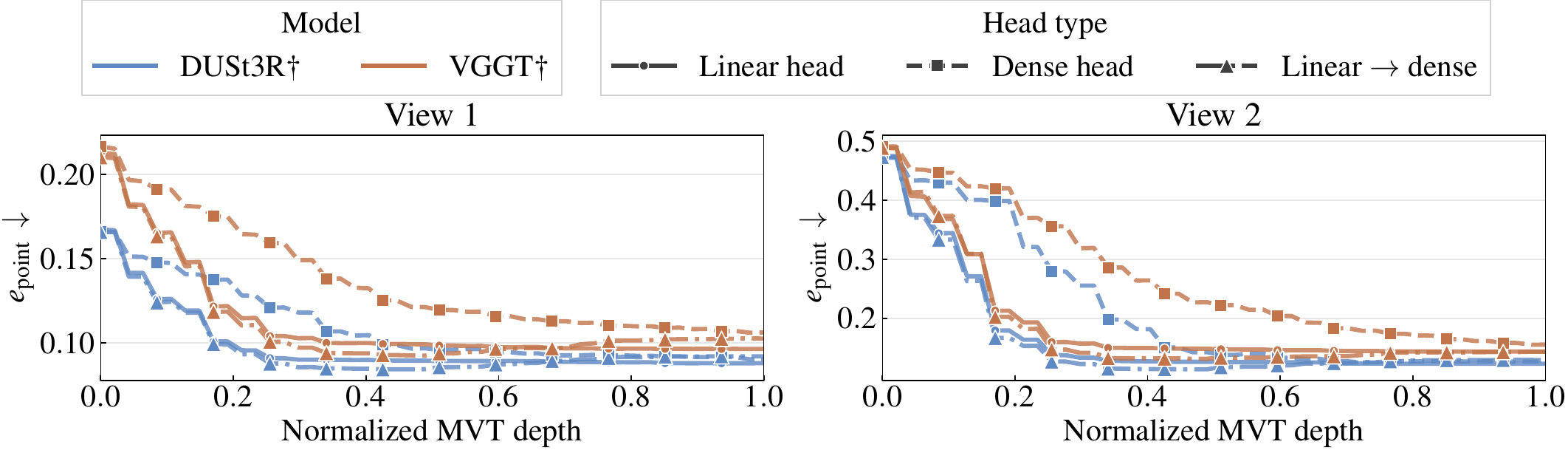}
    \caption{\textbf{Output head design affects reconstruction dynamics.} Probed pointmap error \(e_{\mathrm{point}}\) on HM3D. Linear-head training and the linear-to-dense curriculum yield fast error reduction, particularly in non-reference view~2. Meanwhile, training with a dense head from scratch leads to slower reduction in reconstruction error. \(\dagger\) denotes our re-implementations with MVTs trained from scratch.}
    \label{fig:app-head-training}
\end{figure}

\paragraph{Effect of output head on error dynamics.}\label{subsec:app-output-ablation}
We study how output head design and training curriculum affect layer-wise reconstruction dynamics using the DUSt3R-style and VGGT-style MVTs. The DUSt3R-style variant combines CroCoV2, cross-attention, and binocular MVTs with separate per-view attention weights and output heads. The VGGT-style variant combines DINOv2, global attention, and shared attention weights and output heads. Apart from these differences, both use the same alternating-attention implementation described in~\cref{subsec:app-mvt-training}. We compare training with a linear head applied to the final layer, training from scratch with a dense head that aggregates multiple layers, and a linear-to-dense curriculum (\cref{subsec:app-mvt-training}). In the curriculum, we retain the MVT weights obtained at the end of the first stage, discard the final-layer head, and initialize a fresh multilayer head. We then train for 10 additional epochs with a fresh optimizer and learning-rate schedule, while keeping the encoder frozen. We freeze each of the six trained checkpoints and train probes using the protocol described in the main text. We train MVTs and probes on 100,000 HM3D images at a resolution of \(224\times224\), and evaluate on held-out HM3D scenes.

\Cref{fig:app-head-training} shows that, across both architectures, training with a final-layer head yields a faster reduction in probed pointmap error across MVT depth, especially for the non-reference view. When training with a final-layer head, the error approaches its final value within the first few MVT blocks. In contrast, training with a multilayer head from scratch yields a slower decrease, particularly for the VGGT-style variant. Training first with a final-layer head and then with a multilayer head preserves the rapid error reduction across MVT depth. These results explain the differences in error reduction speed across MVT depth observed between DUSt3R and VGGT/DA3 in~\cref{subsec:state}.

\subsection{Role of Individual Layers}\label{subsec:app-role}

\paragraph{Quantitative analysis.}

We present detailed results for the layer-role analysis in the main text. For each setting, we report three errors: \(e_{\mathrm{point}}\), the per-pixel relative pointmap error defined in~\Cref{eq:app-pointmap-absrel}; \(e_{\mathrm{proc}}\), the Procrustes-aligned variant described in \cref{subsec:app-pointmap_eval_details}; and \(e_{\mathrm{mv}}\), the cross-view error computed between predictions from different views corresponding to the same underlying 3D point. For each metric \(e \in \{e_{\mathrm{point}},~ e_{\mathrm{proc}},~ e_{\mathrm{mv}}\}\), we report the error across layers for all models, and the Accumulated Layer Contribution Percentage aggregated by layer type (\cref{eq:ALCP}), separately for the reference view and the non-reference view(s).

\begin{figure}[p]
    \centering
    \includegraphics[width=\linewidth,height=0.86\textheight,keepaspectratio]{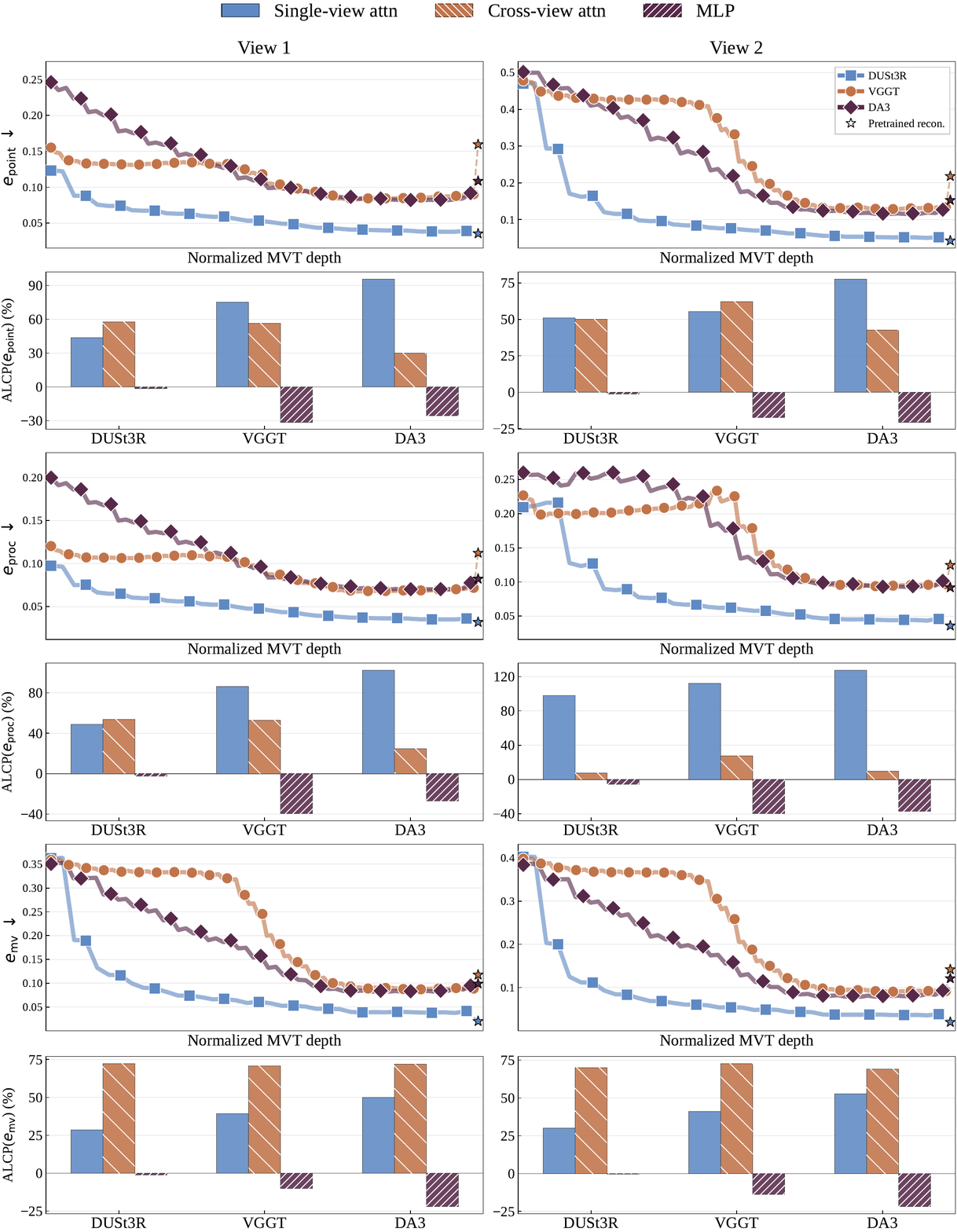}
    \caption{\textbf{Layer-role analysis on HM3D.} Probes for DUSt3R, VGGT, and DA3 are trained and evaluated on HM3D. Rows pair, for each metric, the error across MVT layers for all models with the Accumulated Layer Contribution Percentage aggregated by layer type.}
    \label{fig:app-layer-role-hm3d}
\end{figure}

\begin{figure}[p]
    \centering
    \includegraphics[width=\linewidth,height=0.86\textheight,keepaspectratio]{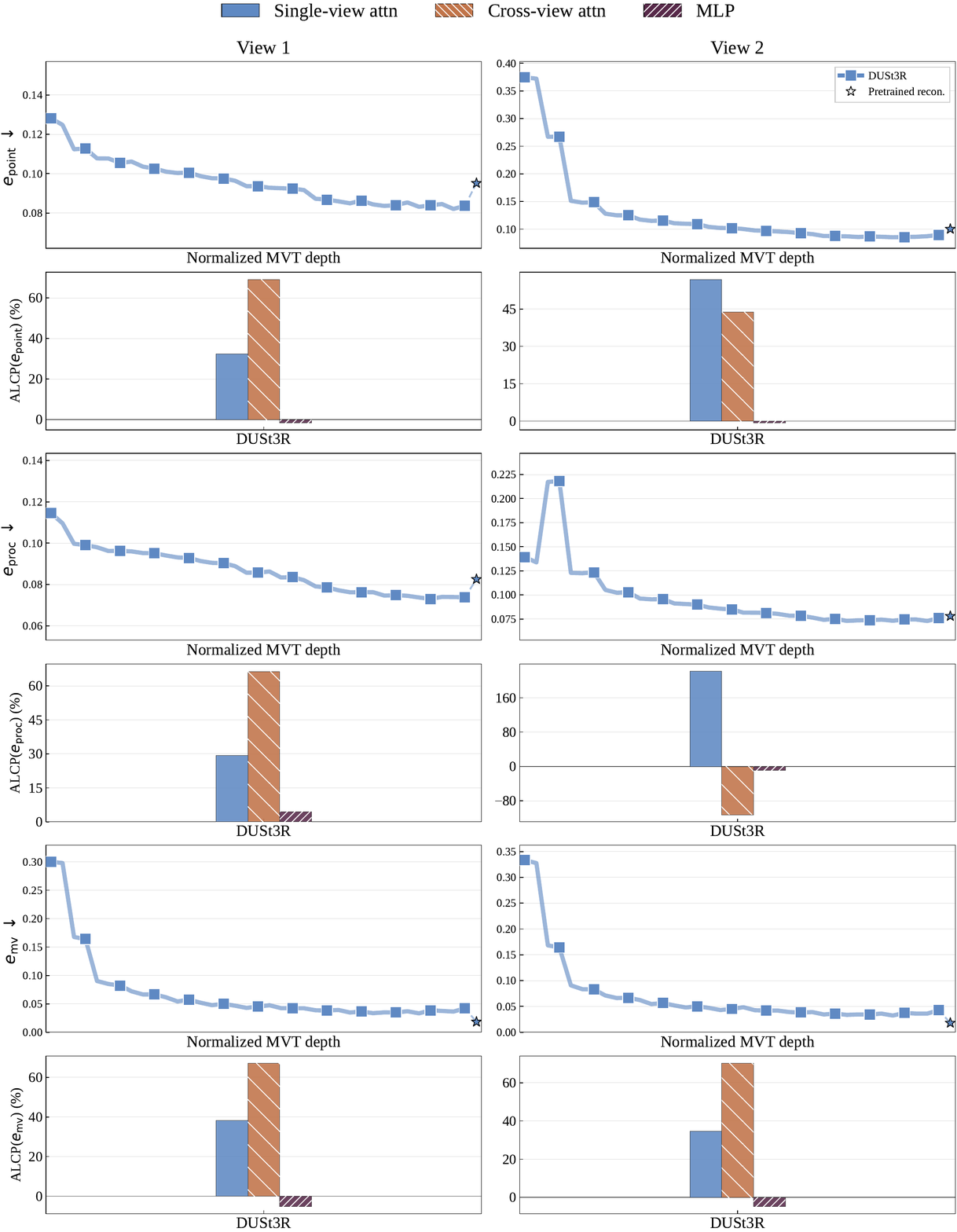}
    \caption{\textbf{Layer-role analysis on BlendedMVS.} DUSt3R probes are trained and evaluated on BlendedMVS. Rows pair, for each metric, the error across MVT layers for DUSt3R with the Accumulated Layer Contribution Percentage aggregated by layer type.}
    \label{fig:app-layer-role-blendedmvs}
\end{figure}

\begin{figure}[p]
    \centering
    \includegraphics[width=\linewidth,height=0.86\textheight,keepaspectratio]{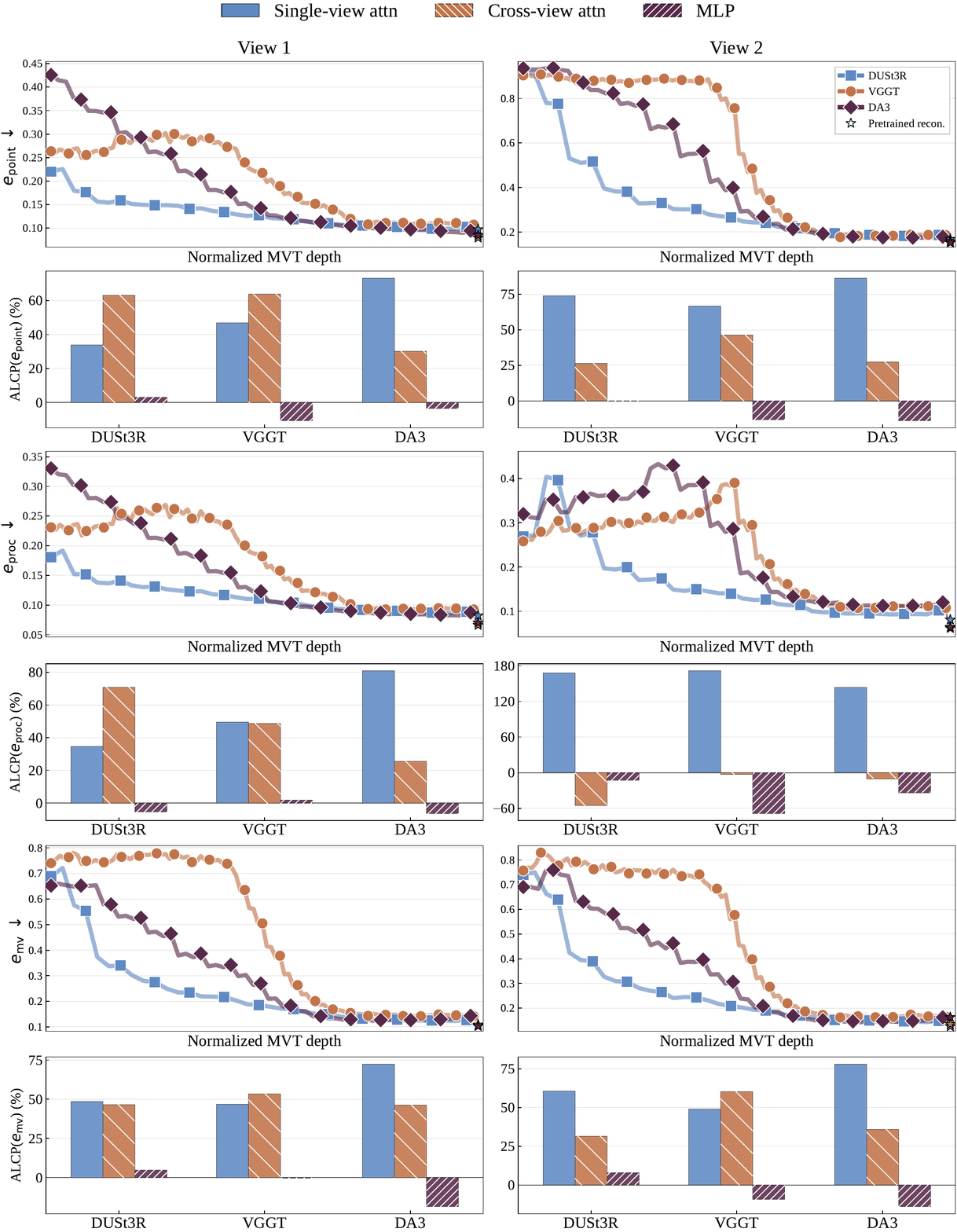}
    \caption{\textbf{Zero-shot layer-role analysis on the MapAnything benchmark.} Probes trained on HM3D are evaluated zero-shot on the MapAnything benchmark datasets (TartanAirv2~\cite{wang2020tartanair}, ScanNet++~\cite{yeshwanthliu2023scannetpp}, and ETH3D~\cite{schoeps2017cvpr}). Rows pair, for each metric, the error across MVT layers for all models with the Accumulated Layer Contribution Percentage aggregated by layer type.}
    \label{fig:app-layer-role-mapanything}
\end{figure}

\begin{figure}[p]
    \centering
    \includegraphics[width=\linewidth,height=0.86\textheight,keepaspectratio]{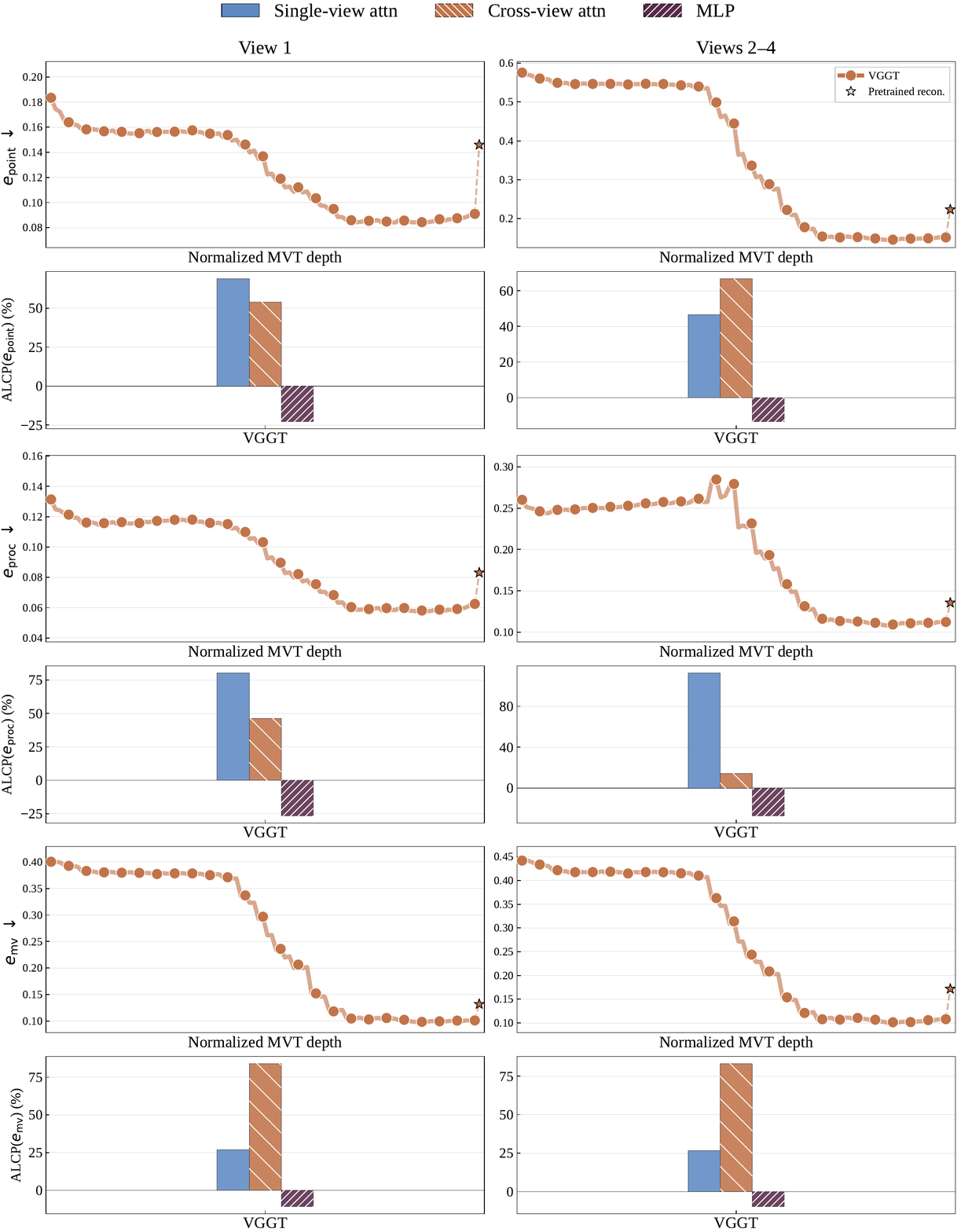}
    \caption{\textbf{Layer-role analysis for four-view VGGT on HM3D.} VGGT probes are trained and evaluated with four HM3D views; the right column averages views~2--4. Rows pair, for each metric, the error across MVT layers for VGGT with the Accumulated Layer Contribution Percentage aggregated by layer type.}
    \label{fig:app-layer-role-vggt-nviews4}
\end{figure}

We use the same layout for four evaluation settings. First, \Cref{fig:app-layer-role-hm3d} reports the main two-view HM3D setting: probes for DUSt3R, VGGT, and DA3 are trained on HM3D and evaluated on held-out HM3D scenes. This is the setting used for the main layer-role claims. Second, \Cref{fig:app-layer-role-blendedmvs} reports DUSt3R probes trained and evaluated on BlendedMVS, which tests whether the same decomposition of layer roles appears on a real multi-view stereo dataset. Third, \Cref{fig:app-layer-role-mapanything} reports the HM3D-trained probes evaluated zero-shot on the MapAnything benchmark datasets. Finally, \Cref{fig:app-layer-role-vggt-nviews4} reports VGGT probes trained and evaluated on HM3D with four input views, where the reference-view probe is applied to view~1 and the non-reference probe is shared across views~2--4. This setting checks whether the two-view observations persist when a feed-forward MVT receives more views.

\begin{table}[!t]
    \centering
    \small
    \caption{\textbf{Attention perturbation in DUSt3R.} Increase in multi-view consistency error \(e_{\mathrm{mv}}\) on HM3D relative to the unperturbed model when the attention maps of a single layer are mixed with a uniform map of strength \(\alpha\). We report the mean and the maximum increase across blocks for each layer type.}
    \label{tab:app-attn-perturbation}
    \begin{tabular}{llrrrrr}
        \toprule
        & \textbf{Layer type} & \(\alpha=0.05\) & \(\alpha=0.1\) & \(\alpha=0.25\) & \(\alpha=0.5\) & \(\alpha=1\) \\
        \midrule
        \multirow{2}{*}{Mean} & Cross-view attention & \(+3.1\%\) & \(+9.1\%\) & \(+31.2\%\) & \(+75.9\%\) & \(+448.4\%\) \\
        & Single-view attention & \(+0.4\%\) & \(+1.4\%\) & \(+6.5\%\) & \(+19.8\%\) & \(+68.3\%\) \\
        \midrule
        \multirow{2}{*}{Max} & Cross-view attention & \(+28.1\%\) & \(+75.0\%\) & \(+227.2\%\) & \(+463.3\%\) & \(+3497.7\%\) \\
        & Single-view attention & \(+2.6\%\) & \(+8.8\%\) & \(+35.4\%\) & \(+92.1\%\) & \(+264.0\%\) \\
        \bottomrule
    \end{tabular}
\end{table}

\paragraph{Attention perturbation.}\label{subsec:app-attn-perturbation}
To further isolate the roles of single-view and cross-view attention in DUSt3R, we perturb one attention layer at a time during inference. Each attention map of the perturbed layer is mixed with a uniform attention map with strength \(\alpha\): \(\alpha=0\) leaves the model unchanged, while \(\alpha=1\) fully replaces the attention map and removes the layer's ability to select tokens. We then measure the increase in multi-view consistency error \(e_{\mathrm{mv}}\) on HM3D relative to the unperturbed model. \Cref{tab:app-attn-perturbation} reports this increase averaged across blocks and its maximum increase.

Across all perturbation strengths, perturbing cross-view attention causes substantially greater degradation than perturbing single-view attention, with effects \(5.03\) to \(13.25\) times larger on average. Even a small perturbation of \(\alpha=0.05\) increases the error by up to \(28.1\%\) for a cross-view layer, compared with at most \(2.6\%\) for a single-view layer. At larger perturbations, the most influential cross-view layer is in block~0, consistent with our probing results identifying it as important for multi-view alignment. These interventions provide evidence that multi-view consistency depends on selective cross-view attention: disrupting its ability to retrieve informative tokens from the other view increases \(e_{\mathrm{mv}}\) far more than an equivalent single-view attention perturbation.

\paragraph{Effect of MLP layers on reconstruction.}\label{subsec:app-mlp-ablation}
We test whether the MVT MLPs are necessary for reconstruction using the VGGT-style architecture described in~\cref{subsec:app-mvt-training}. We compare this baseline with a variant that removes the MLP layers after both single-view and cross-view attention while retaining the same depth and width. Excluding the frozen encoder and reconstruction head, this reduces the parameter count from \(302.4\)M to \(100.8\)M. We additionally train an MLP-free variant that matches the baseline parameter budget approximately by increasing MVT depth from 12 to 24 MVT blocks and embedding dimension from 1024 to 1216 channels. All three models are trained from scratch on 100,000 HM3D images at \(224\times224\) resolution with the setup described in~\cref{subsec:app-mvt-training}. We evaluate their output 3D reconstructions on held-out HM3D scenes.

\Cref{tab:app-mlp-ablation} shows that removing MLPs at fixed depth and width increases pointmap error by only \(14.6\%\) while using one-third as many trainable parameters. Increasing the depth and width of the MLP-free model to approximate the baseline parameter count improves error by \(5.6\%\) relative to the baseline. This suggests that reallocating MLP capacity to attention can improve reconstruction in this small-scale setting.

\makeatletter
\setlength{\@fptop}{0pt}
\makeatother

\clearpage

\makeatletter
\setlength{\@fptop}{0pt plus 1fil}
\makeatother

\paragraph{Qualitative analysis.}\label{subsec:app-role-viz}

\begin{figure}[!t]
    \centering
    \includegraphics[width=1\linewidth]{figures/main/low_res/sa_ca.pdf}
    \caption{\textbf{Probed pointmaps in DUSt3R early MVT blocks.} We visualize pointmaps predicted after the first single-view and cross-view attention operations. Results are shown both before and after Procrustes alignment. Points from view~1 are colored in green, and points from view~2 in red. Cross-view attention aligns the two views but disrupts the geometric consistency within view~2. Subsequent single-view attention restores intra-view geometric structure.}
    \label{fig:app-layer-role}
    \label{fig:layer-role-viz}
\end{figure}

In \Cref{fig:app-layer-role}, we illustrate the role of cross-view and single-view attention with a visualization of an example probed from DUSt3R. We show intermediate reconstructions from early MVT layers, starting before any cross-view communication occurs. For each layer, we display two versions of the predicted pointmap: the raw probe output and the version after Procrustes alignment with the ground truth. As described in \cref{subsec:pointmap-eval}, this alignment procedure allows us to evaluate the geometric consistency of the second view independently of the relative pose error introduced by imperfect alignment with the first view during early reconstruction stages. We also report the L2 error before Procrustes alignment for each view with respect to the ground truth for each pointmap sequence.

The initial reconstruction of the first view is already accurate and exhibits low error. Probe outputs indicate that this reconstruction remains largely unchanged across subsequent layers and effectively acts as a stable anchor. In contrast, the initial reconstruction of the second view is not aligned with the coordinate system of the first view. However, the Procrustes-aligned sequence reveals that the second-view pointmap already captures the scene geometry (e.g., the rigidity of the wall). The first cross-view attention layer then registers the second view with the first, substantially reducing the error in the direct probe output. Examining the same transformation in the Procrustes-aligned sequence shows that it temporarily disrupts geometric consistency in the second view (for instance, the rigidity of the wall) resulting in a sharp increase in Procrustes-aligned error. The subsequent single-view attention operation restores the internal geometric structure of the second view, leading to a decrease in Procrustes-aligned error.

These observations illustrate the complementary roles played by the two attention mechanisms. Cross-view attention aligns the second view with the first through non-rigid updates, while single-view attention subsequently improves internal geometric consistency.

\subsection{Role of correspondences}\label{subsec:app-corresp-result}

\begin{figure}[h]
    \centering
    \includegraphics[width=0.96\linewidth]{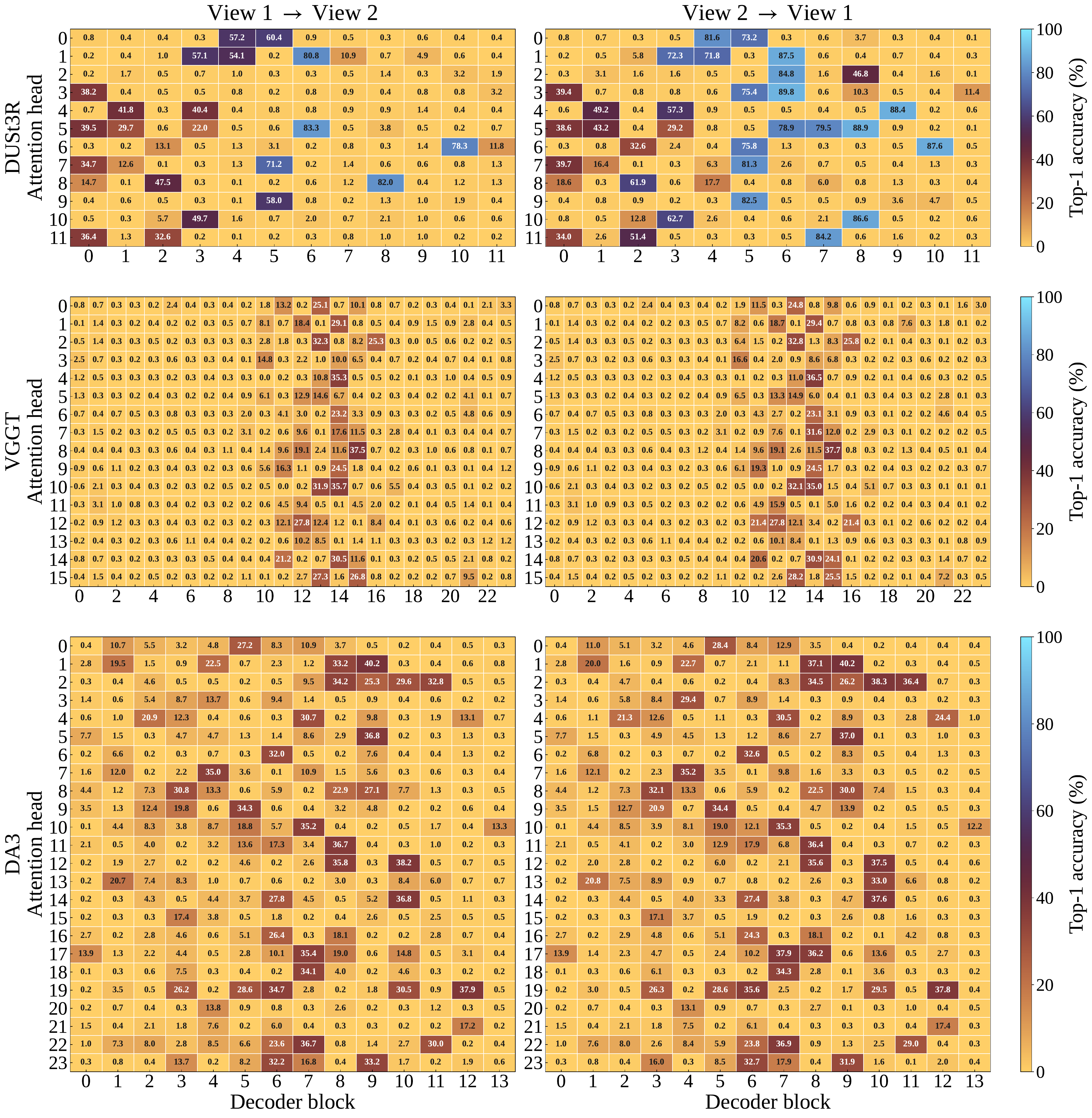}
    \caption{\textbf{Per-head correspondence accuracy.} Top-1 correspondence accuracy (\%) of every cross-view attention head on HM3D, for both attention directions. Correspondences improve across MVT blocks until the middle of the MVT for all models. DUSt3R presents the highest accuracy across all models.}
    \label{fig:corresp_tables}
\end{figure}

\begin{figure}[h]
    \centering
    \includegraphics[width=1\linewidth]{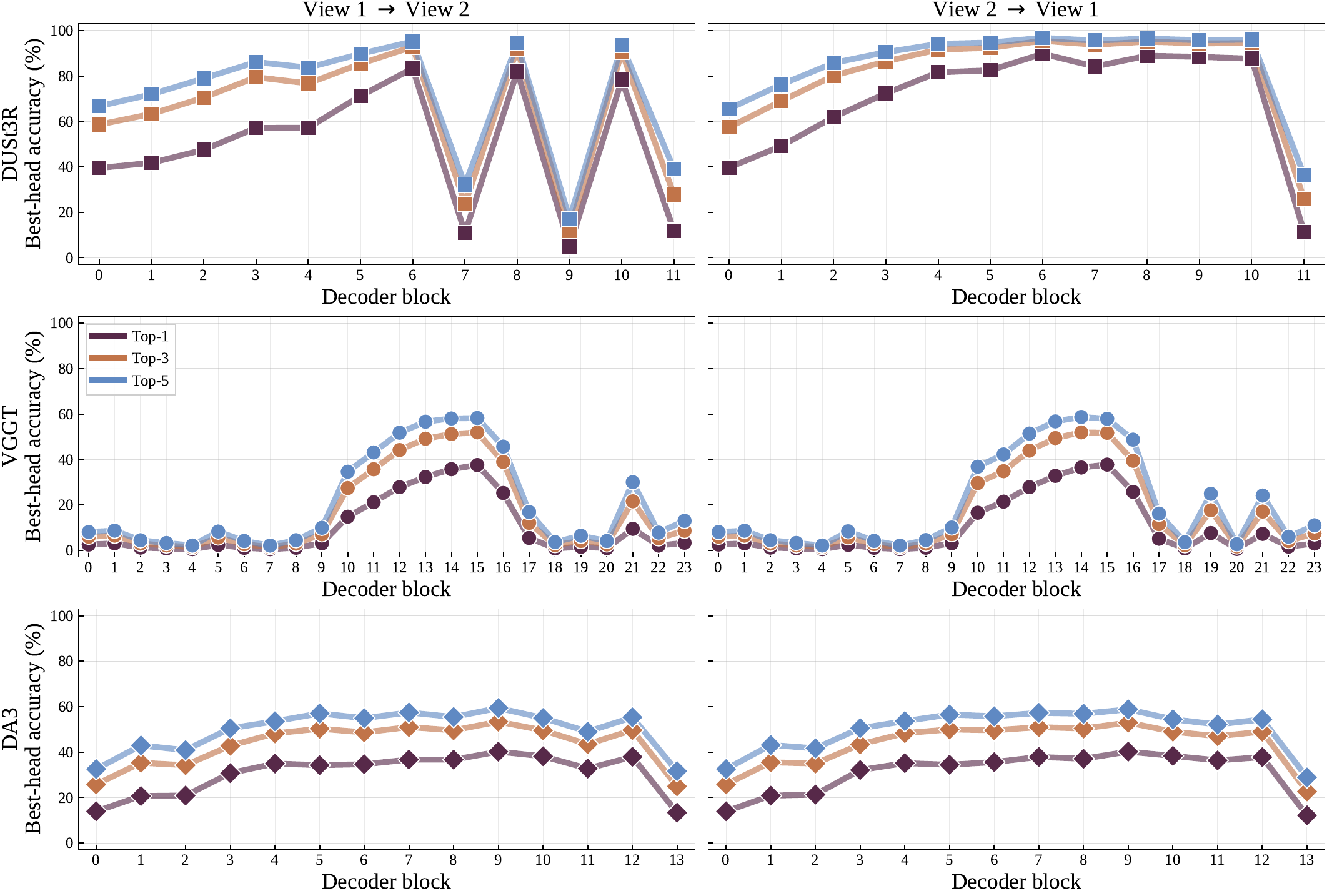}
    \caption{\textbf{Best correspondence accuracy.} Top-1, top-3, and top-5 correspondence accuracy of the best cross-view attention head in each MVT block on HM3D, for both attention directions.}
    \label{fig:corresp_best}
\end{figure}

\paragraph{Per-head correspondences.} For every model, we report the per-head correspondence accuracy in \Cref{fig:corresp_tables}, following the methodology described in the main text. Accuracy is aggregated over all valid query patches in the evaluation set. We find that correspondences for the second view improve across the first half of the MVT for all models. In particular, the correspondence accuracy is significantly higher for DUSt3R than for other models. We confirm this trend and report the best-performing head in \Cref{fig:corresp_best} for varying top-k thresholds.

\paragraph{Encoder ablation.} To understand why DUSt3R yields much stronger correspondences than VGGT and DA3, we measure correspondence accuracy in the MVTs of our design space ablation (\cref{subsec:mvt_ablations}), trained from scratch with the shared architecture described in \cref{subsec:app-mvt-training}. We vary the frozen pretrained encoder (CroCoV2 or DINOv2) and the cross-view attention mechanism (cross-attention or global attention), and keep all other components fixed. For scene-level evaluation, we train on HM3D and evaluate on held-out scenes. For object-centric evaluation, we train on DTC and evaluate on held-out DTC objects and on GSO objects. We report the top-1 View~2 \(\rightarrow\) View~1 accuracy of the best-performing head, using the zero-shot correspondence estimator of the main text. \Cref{tab:corresp_mvt_ablation} shows that the encoder has the largest effect: with cross-attention, replacing DINOv2 with CroCoV2 raises accuracy from 40.1\% to 73.9\% on HM3D, and by about 20 points on both object-centric datasets. The attention mechanism has a smaller effect. With CroCoV2, cross-attention outperforms global attention on all datasets. With DINOv2, global attention is slightly better on HM3D but much worse on object-centric data. Overall, the DUSt3R-like configuration (CroCoV2 with cross-attention) achieves the highest accuracy on every dataset, which suggests that the cross-view matching objective of CroCo pretraining favors the emergence of correspondences in cross-view attention.
\begin{table}[!t]

    \centering
    \small
    \caption{\textbf{MVT correspondence ablation.} Top-1 View~2 $\rightarrow$ View~1 accuracy of the best-performing attention head. CA and GA denote cross- and global attention.}
    \label{tab:corresp_mvt_ablation}
    \setlength{\tabcolsep}{3pt}
    \begin{tabular}{@{}lccc@{}}
        \toprule
        & \multicolumn{1}{c}{\textbf{Scene-level}} & \multicolumn{2}{c}{\textbf{Object-level}} \\
        \cmidrule(lr){2-2}\cmidrule(lr){3-4}
        \textbf{Model} & \textbf{HM3D} & \textbf{GSO} & \textbf{DTC} \\
        \midrule
        CroCoV2 + CA & \textbf{73.93\%} & \textbf{55.53\%} & \textbf{52.74\%} \\
        CroCoV2 + GA & 67.92\% & 50.48\% & 50.13\% \\
        DINOv2 + CA  & 40.06\% & 35.45\% & 32.54\% \\
        DINOv2 + GA  & 43.40\% & 21.89\% & 21.97\% \\
        \bottomrule
    \end{tabular}
\end{table}

\begin{figure}[ht]
    \centering
    \includegraphics[width=1\linewidth]{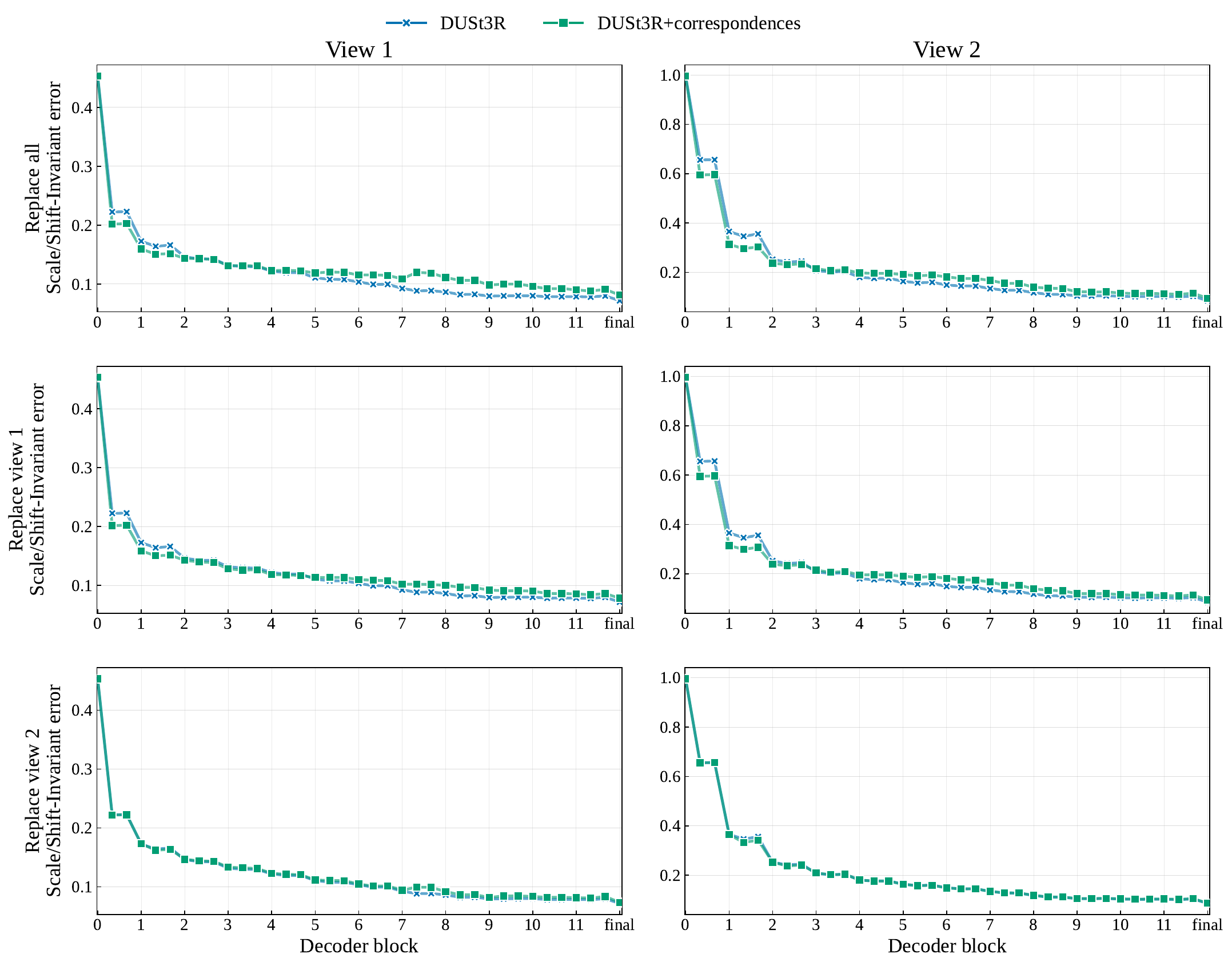}
    \caption{\textbf{Replacing cross-view attention with correspondences.} We replace DUSt3R cross-view attention maps with ground-truth correspondences for heads that perform correspondence search.}
    \label{fig:dust3r_gt_attn}
\end{figure}

\paragraph{Replacing attention heads with GT correspondence maps.} In DUSt3R, we replace cross-view attention maps by the ground-truth correspondence map for patch tokens with a valid correspondence in the other view. We apply this only for attention heads that perform correspondence search. To automatically find these heads, we use the per-head correspondence accuracy in~\cref{fig:corresp_tables} and modify the heads whose accuracy is higher than a threshold. We choose 2\% as the threshold. We then evaluate the modified model and report the results compared to the original DUSt3R (\cref{fig:dust3r_gt_attn}). We explore replacing heads in both view streams, as well as in each view stream separately. We find that replacing the heads in either view stream separately does not affect the result significantly. However, a slight decrease in performance is observed when replacing both.

\subsection{Additional Visualizations}\label{subsec:app-visualizations}

\paragraph{Pointmap visualizations.} Additional pointmap visualizations for early MVT blocks are provided in \Cref{fig:app:other_refinement}, along with colored refinement examples in \Cref{fig:app_others}.

\begin{figure*}[ht]
    \centering
    \includegraphics[width=1\linewidth]{figures/appendix/low_res/Appendix_other_examples.pdf}
    \caption{\textbf{Additional pointmap visualizations.} The two simpler top scenes (room and corridor) converge after a single cross- and self-attention iteration. The four more challenging bottom cases (bicycle, couch corner, chair with Lego, counter) require 3 to 6 iterations to correctly align the second-view patches (red). In the final opposing-view example (stump), reconstruction fails, likely due to early convergence to a local minimum.} 
    \label{fig:app:other_refinement}
\end{figure*}
\begin{figure*}[ht]
    \centering
    \includegraphics[width=1\linewidth]{figures/appendix/low_res/Appendix_other_examples_2.pdf}
    \caption{\textbf{Additional visual examples with RGB colors}.}
    \label{fig:app_others}
\end{figure*}

\begin{figure}[t]
  \centering
  \begin{minipage}[t]{0.48\linewidth}
    \includegraphics[width=\linewidth]{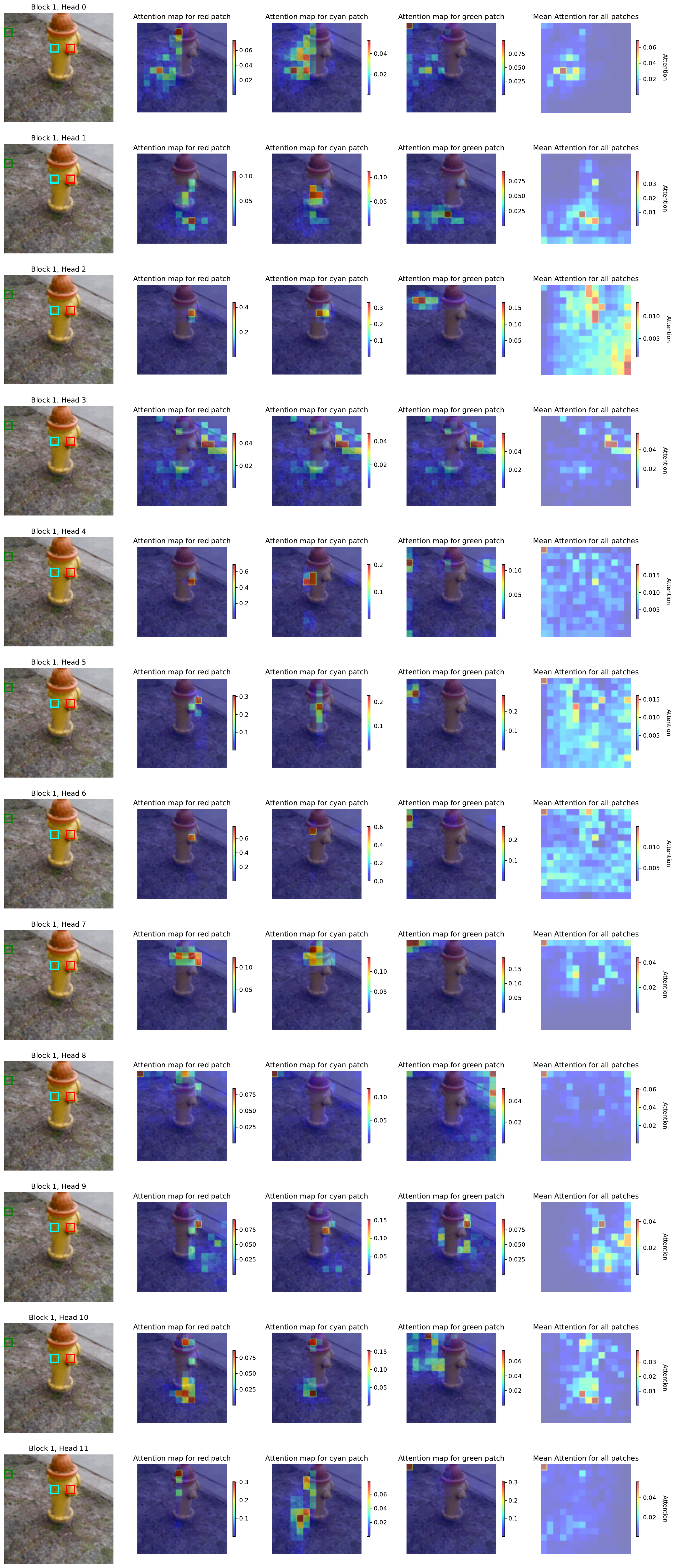}
    \caption{\textbf{Visualization of DUSt3R self-attention maps} in the second MVT block of the second view. We visualize the attention maps for three query patches shown on the query image. Heads 1, 2, 4, 5, 6, and 11 mainly attend to local patches around the query; heads 0, 3, 8, and 9 are identified as attending to \textit{register tokens}. The locations of these registers vary across examples.}
    \label{fig:sa_hydrant}
  \end{minipage}\hfill
  \begin{minipage}[t]{0.48\linewidth}
    \includegraphics[width=\linewidth]{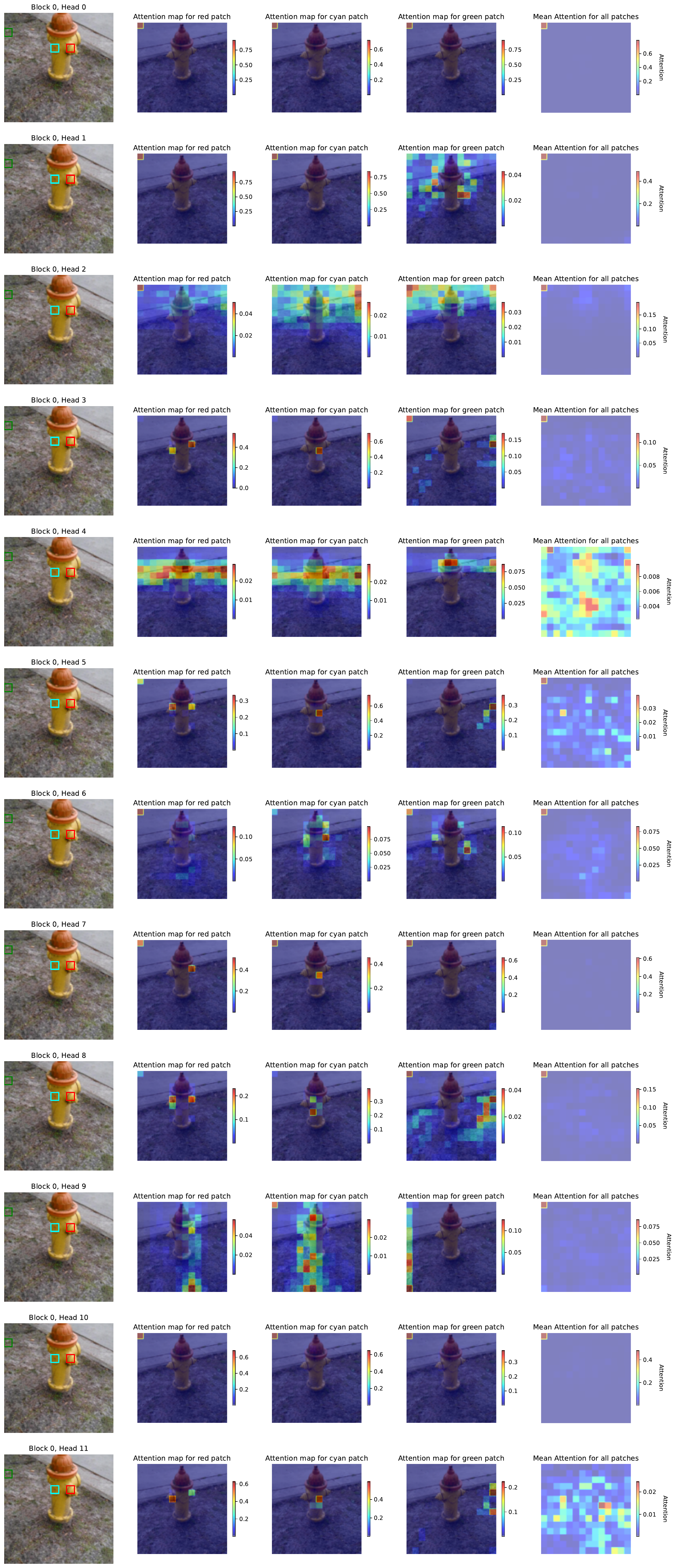}
    \caption{\textbf{Visualization of DUSt3R cross-attention maps} in the first MVT block of the second view. We visualize the attention maps for three query patches shown on the query image. Heads 3, 5, 7, 8, and 11 act as correspondence heads, while heads 0 and 10 consistently attend to the top-left \textit{register token}. Heads 9 and 4 exhibit spatial behavior, performing vertical and horizontal searches, respectively.}
    \label{fig:ca_hydrant}
  \end{minipage}
\end{figure}

\paragraph{Self-attention maps visualizations.}\label{app:sa}
We visualize self-attention maps in the second MVT block of the second view for the \texttt{hydrant} example in \Cref{fig:sa_hydrant}. We selected this layer because it contributed most to reducing the second-view pointmap error after Procrustes alignment. We identify heads that mainly attend to neighboring patches of the query and others that consistently attend to a fixed subset of patch tokens across all queries. We refer to these heads as \textit{register heads} and these subsets of tokens as \textit{register tokens}, which likely correspond to the high-norm tokens identified in \cite{darcet2024vision}. Note that our use of this term differs from their paper, where it refers to additional tokens introduced to overcome this phenomenon. This behavior repeats across samples and can also be observed in other MVT blocks; however, we show only one block on a single example for brevity.

\paragraph{Cross-attention maps visualizations.}\label{app:ca}
We visualize cross-attention maps in the first MVT block for the \texttt{hydrant} example in \Cref{fig:ca_hydrant}. Similar to self-attention, the attention maps indicate a specialization of individual heads for particular tasks. Noticeably, heads 3, 5, 7, 8, and 11 attend to candidate corresponding patches for the query, whenever correspondences exist. The behavior of these correspondence heads is consistent across samples. We also observe heads attending to \textit{register tokens}, as in the self-attention map. Besides, we notice that cross attention maps tend to be significantly sharper than self-attention ones.

\end{document}